\documentclass[sn-basic]{sn-jnl}

\usepackage{graphicx}%
\usepackage{multirow}%
\usepackage{amsmath,amssymb,amsfonts}%
\usepackage{amsthm}%
\usepackage{mathrsfs}%
\usepackage[title]{appendix}%
\usepackage{xcolor}%
\usepackage{textcomp}%
\usepackage{manyfoot}%
\usepackage{booktabs}%
\usepackage{algorithm}%
\usepackage{algpseudocode}%
\usepackage{listings}%
\usepackage{todonotes}
\usepackage{xcolor}
\definecolor{darkgreen}{RGB}{0, 100, 0}

\newcommand{\new}[1]{\textcolor{black}{#1}}
\newcommand{\framework}{\textsc{SimHawNet}}
\newcommand{\us}{\textsc{SimHawNet }}
\newcommand{\fram}{\textsc{HawNet}}

\newcommand{\ind}{\mathbbm{1}}

\newcommand{\Hcal}{\mathcal{H}}
\newcommand{\Hcalt}{\mathcal{H}_t}

\newcommand{\Ecal}{\mathcal{E}}

\newcommand{\vx}{\bm{x}}

\newcommand{\ptitle}[1]{\textbf{#1.}}
\newcommand{\ptitleskip}[1]{\vspace{4pt}\textbf{#1.}}

\usepackage{tkz-berge}
\usepackage{bm}
\usepackage{bbm}
\usepackage{xcolor}%
\usepackage{subcaption}
\usepackage{pifont}

\theoremstyle{thmstyleone}%
\newtheorem{theorem}{Theorem}
\newtheorem{proposition}[theorem]{Proposition}%

\theoremstyle{thmstyletwo}%

\theoremstyle{thmstylethree}%
\newtheorem{definition}{Definition}%

\begin{document}

\title[\framework]{\framework: A Modified Hawkes Process for Temporal Network Simulation}


\author[1]{\fnm{Mathilde} \sur{Perez}}\email{mathilde.perez@telecom-paris.fr}
\equalcont{These authors contributed equally to this work.}
\author[2]{\fnm{Raphaël} \sur{Romero}}\email{raphael.romero@ugent.be}
\equalcont{These authors contributed equally to this work.}
\author[2]{\fnm{Bo} \sur{Kang}}
\author[2]{\fnm{Tijl} \sur{De Bie}}
\author[2]{\fnm{Jefrey} \sur{Lijffijt}}
\author[1]{\fnm{Charlotte} \sur{Laclau}}
\affil[1]{LTCI, Télécom Paris,
Institut Polytechnique de Paris, France}
\affil[2]{IDLab, Ghent University, Belgium}


\abstract{Temporal networks allow representing connections between objects while incorporating the temporal dimension. While static network models can capture unchanging topological regularities, they often fail to model the effects associated with the causal generative process of the network that occurs in time. Hence, exploiting the temporal aspect of networks has been the focus of many recent studies.
In this context, we propose a new framework for generative models of continuous-time temporal networks. We assume that the activation of the edges in a temporal network is driven by a specified temporal point process. This approach allows to directly model the waiting time between events while incorporating  time-varying history-based features as covariates in the predictions. Coupled with a thinning algorithm designed for the simulation of point processes, \framework\ enables simulation of the evolution of temporal networks in continuous time. Finally, we introduce a comprehensive evaluation framework to assess the performance of such an approach, in which we demonstrate that \framework\ successfully simulates the evolution of networks with very different generative processes and achieves performance comparable to the state of the art, while being significantly faster.
}

\keywords{Temporal Networks, Temporal Point Process, Graph Simulation}



\maketitle

\section{Introduction}\label{sec:introduction}

\ptitle{Motivation} The study of continuous-time temporal networks has attracted growing interest in the past decades, as time-stamped relational data emerges from various sources, including e-mail communication, e-commerce, and even neuron firing data \citep{Goldenberg2009,Holme2012}. 
Understanding how networks evolve over time provides valuable insights into the underlying mechanisms driving their behavior. It also enables the development and testing of methods for analyzing and predicting real-world (social) networks' dynamics. Examples of use cases include the study of opinion dynamics, epidemic spreading, and recommender systems \citep{panzarasa2009patterns, machens2013infectious}. 

Attempts to model such networks have traditionally relied on a discretization of time, allowing the representation of the temporal network as a sequence of static graphs\footnote{Note that we use the terms \textit{network} and \textit{graph} interchangeably and more specifically the terms \textit{temporal network} and \textit{dynamic graph} also have the same meaning here and in existing literature.} \citep{Krivitsky2014,Kim2018,Durante2016}.
As described by \cite{Latapy2017}, these attempts to read this problem through the well-documented lens of graph theory fail to capture mechanisms coming from the causal nature of temporal networks. 

Moreover, temporal relational data is often available with a high level of time granularity, allowing us to consider the time stamps as continuous \citep{Rozenshtein2019}. The interaction patterns are often heavily localized in time, exhibiting specific properties such as \textit{burstiness} \citep{NaokiMasuda}. For instance, a good model for temporal social networks should be able to give an idea of the duration that will separate two interactions between users, given a certain historical context.
To this date, only a few tools are available for continuous-time temporal network analysis, e.g., methods from computational social sciences \citep{Butts2008} and point process modeling \citep{Bacry2018}. 


To adapt these tools for temporal graph simulation, we propose incorporating features from computational social sciences into a temporal point process model. This approach elegantly and flexibly introduces inductive bias, enabling the assessment of the statistical properties of an observed temporal graph. The inter-event times are modeled using modified Hawkes processes, conditioned on both the structural nature through node embeddings and the temporal dynamics through time-varying features of the observed interactions within the graph.
Second, we leverage the resulting model to simulate dynamic graphs that mimic the temporal evolution and interactions within the network. Our approach opens the avenue for several applications including evaluating algorithms for network analysis, studying the effects of interventions or policy changes (e.g. fairness constraints), or training and testing machine learning models for tasks such as link prediction or community detection.

\ptitleskip{Contributions}
This work aims to introduce a novel framework that allows the translation of temporal point processes, specifically Hawkes processes, on a generative temporal network model. To this end, before presenting the framework, we introduce the necessary notation and concepts in Section~\ref{sec:background} and present the expression of the intensity function that we consider in this work. Based on this expression, in Section~\ref{sec:approach}, we introduce the Hawkes network model (\fram) that includes modeling assumptions for the inter-arrival times between events for each edge, to measure the salience of temporal network properties in observed data. We show how to perform Maximum Likelihood Estimation of model parameters on an observed network.
In Section~\ref{sec:sim}, we introduce the simulation solution, that is built upon the learned model. The proposed algorithm utilizes a thinning algorithm inspired by \cite{Ogata1981} originally proposed to simulate point processes. The overall framework is illustrated in Figure~\ref{fig:framework} and referred to as \us. Finally, in Section \ref{sec:xp}, we develop an evaluation framework comprising dynamic synthetic graphs derived from established static simulation processes, alongside suitable metrics for assessing the efficacy of a dynamic graph simulation model. We then demonstrate \us's capability to accurately replicate the behavior of dynamic graphs both within this framework and in a real-world application. The full code is available online\footnote{\label{ftn:code}Code available at: \url{https://github.com/MathouP31/SimHawNet}}. Conclusions and perspectives of extensions are presented in Section~\ref{sec:conclusion}. 


\section{Related Work}
To the best of our knowledge, the previous work closest to the method introduced in this paper is presented by \cite{huangMutuallyExcitingLatent2022}, who propose to model relational events between nodes using mutually-exciting Hawkes processes, where the baseline intensity is influenced by node proximity in a latent space and by node-specific sender and receiver effects. However, they rely on static node effect terms to define the dynamic part of conditional intensity functions, while in our case we rely on time-varying features to model the temporal evolution of the intensity.

\ptitle{Point Process Modeling of Temporal Networks} 
More generally, Point Process modeling of Temporal Networks is a popular paradigm encompassing many previous studies. Seminal works coming from the computational social sciences and machine learning community propose to model temporal networks through their intensity functions to assess the effect of the temporal network topology on the rates of link formation \citep{Butts2008,DuBois2013,Perry2013e, Farajtabar2018,Vu2011}. Specifically, relational event modeling also relies on a piece-wise constant intensity hypothesis to define the likelihood of the observed network. In our case, we provide greater flexibility by allowing the insertion of a customized hazard function on the edge-inter-event times. This function can be made dependent or independent of the evolution of other edges in the network, depending on specific requirements. In the same spirit, \cite{Passino2021} propose to posit a mutually exciting model between the different edges, such that the interactions occurring for a given edge influence the ones occurring on neighboring edges. However, they rely on time-varying node embeddings to define the conditional intensity functions and focus on parameter estimation rather than generation.


\ptitle{Dynamic Graph Embedding}
Dynamic graph embedding methods such as \cite{Dai2016,trivedi2018dyrep,10281384,Rastelli_Corneli_2023} also model the interaction times between edges using point processes. More precisely, they use the log-likelihood of the model to \emph{learn time-varying node embeddings}. In the same direction, learning time-varying embeddings (not necessarily by directly using point process models) from a history of past interaction is a task attracting growing interest \citep{Rossi2020, xuInductiveRepresentationLearning2020b, Kumar2019}. In our case, however, static embeddings are used in the base rate but we separately model the influence of past interactions on the rate and do not require the embeddings to evolve through time.

\ptitle{Dynamic Stochastic Block Modeling}
Furthermore, extensions of the Stochastic Block Model (SBM) to the dynamic case have also considered inhomogeneous Poisson processes to model the conditional interaction between pairs of nodes \citep{CorneliLR16,matias2018, arastuieCHIPHawkesProcess2020}. However, these methods usually assume that some latent groups in the network drive the intensity associated with the pairs, that they aim to recover.  \\

\textit{Remark on the connection with graph generation}: it is important to highlight the significant distinction between dynamic graph simulation and (non-)autoregressive graph generation methods \citep{Bonifati2020}. While graph generation models primarily concentrate on generating static graphs, they typically only address the presence of edges without considering the possibility of edge activation multiple times. Moreover, an essential difference lies in the disregard of the temporal order by generative models, leading to an arbitrary order in the creation of edges.


\section{Temporal Networks and Hawkes process}
\label{sec:background}
This section introduces the used notation and the fundamental concept underlying our framework, namely temporal point processes. We then demonstrate how to design a first-order Markov chain model for a temporal network, based on a modified Hawkes model for the edge inter-arrival times.

\subsection{Notation}
Let $\mathcal{U}$ and $\mathcal{V}$ denote a set of source and destination nodes respectively, $\mathcal{E} \subset \mathcal{U} \times \mathcal{V}$ the set of possible edges.
Temporal network data takes the form of a sequence of time-stamped network events $\mathcal{H} = \{(e_m, t_m)\ |\ m=1,..., M\}$, where $ M$ is the number of events, $0<t_1<...<t_M$ is an ordered sequence of positive time stamps, $e_m=(u_m,v_m)\in \mathcal{E}$ is the edge involved in the $m$-th event, and $u_m\in \mathcal{U}$ and  $v_m\in \mathcal{V}$ are the source and destination nodes respectively. For instance, each event $((u,v),t)$ can represent a message sent by user $u$ to user $v$ at time $t$.
For any time $t$ we will denote $\mathcal{H}_t=\{(e_m,t_m) \in \mathcal{H}\ |\ t_m<t\}$ the history of the network up to time $t$. Moreover, for any edge $e =(u,v)\in \mathcal{E}$, we denote $\mathcal{H}^{(e)}=\{(e_m,t_m)  \in \mathcal{H}\ |\ e_m=e\}$ the history of events involving the edge $e$.
Finally, we denote by $\mathcal{H}_t^{(e)}$ the history of events involving the edge $e$, up to time $t$. In our case, we work on undirected graphs, which means $(u,v) = (v,u)$.

\subsection{Temporal Point Processes}
Temporal point processes are stochastic variables whose realizations are sets of positive real-valued time stamps in some time interval $[0, T]$. These variables in our case, can be represented as an event sequence of activation of edges $\mathcal{H}_t = \{ (e_1,t_1), ..., (e_M,t_M)\}$ where the size of this sequence $M$ is a random variable too.

A fundamental quantity in the study of temporal point processes is the \emph{conditional intensity function}, denoted $t\mapsto\lambda(t|\Hcal_t)$, such that for any $t\in[0, +\infty[$, $\lambda(t|\Hcal_t)dt$ measures the average number of events occurring in the interval $[t,t+dt[$, given the event history up to time $t$ \citep{Daley2010}.

A well-known example is the Poisson process, his conditional intensity function is independent of the history, meaning that it is a \emph{deterministic} function: $\lambda(t|\Hcal_t)=\lambda(t)$.
In contrast, \emph{self-exciting} processes are governed by a \emph{stochastic} conditional intensity function, whose value at time $t$ depends on the samples previous to $t$.
While Poisson Processes can be accurate when we have a clear a priori idea of the form of the intensity function, they are unable to adapt to new events. On the contrary self-exciting processes allow the conditional intensity function to take into account past events where historical events have a positive exciting influence on later events. A common self-exciting point process is the Hawkes process \citep{Hawkes1971}, which has a conditional intensity function :
$$\lambda(t| \mathcal{H}_t) = \mu(t) + \sum_{t_{i}<t} \gamma(t-t_i).$$
Where $\mu(t) > 0$  represents a base intensity, $\gamma$ an excitation function, and $t_i \in \mathcal{H}_t$ a previous event.


\subsection{From Hawkes process to first-order Markov chain}
\label{markov}


This section will explain how to transition from a Hawkes process to a Markov chain framework by modeling the conditional intensity function using time-varying features, rather than relying on the entire history of events. This approach enables a more structured representation of the inter-arrival times of events in a temporal network.

To give an expression of the intensity function, we suppose that in the absence of any event, the point process is generated by a deterministic $\lambda(t|\Hcal_t=\emptyset)=\mu(t)$.
We also suppose the existence of a vector-valued stochastic function $t\mapsto \vx(t)$, defined for any time $t\in [0,+\infty[$, called \emph{time-varying features}. Then, given a non-empty history of previous events $\{t_1,...,t_n\}$, we suppose that $\vx(t_n )$ is a sufficient statistic of the history up to time $t_n$, for the prediction of the next event :
$\lambda(t|\Hcal_t=\{t_1,...,t_n\})=\lambda(t|\vx(t_n))$.

Alternatively, the occurrence of each new event can be seen as incrementing the conditional intensity function with a time-varying value.

\begin{definition}
	We define on $[0, +\infty[$ the \emph{intensity increment function}:
	$$t \mapsto\gamma(t|\Hcalt) = h(t-t_n|\vx(t_n)) + \mu(t)$$
\end{definition}
While $h$ is a deterministic function of the time elapsed since the last event, $\gamma$ is a stochastic function of the absolute time of the process so $\gamma$ depends on the history $\Hcalt$ only through the time-varying feature vector $\vx(t_n)$


Conditioned on the feature function $\vx$, the procedure describes a self-exciting process, where the intensity function gets set to the value of the hazard function at every new event.

\begin{proposition}
	The conditional intensity function of the point process defined above is given for any $t\in [0, +\infty[$ by
 \begin{equation}\label{eq:hawnet}
\lambda(t|\Hcalt) = \mu(t) + \sum\limits_{t_n<t}\ind_{t\in [t_{n}, t_{n+1})}\gamma(t|\vx(t_n))
\end{equation}
	where $\ind_{t\in [t_{n}, t_{n+1})}$  is the indicator function asserting that the current time is before the next event.
\end{proposition}

To summarize, this formulation derives from Hawkes processes \citep{Hawkes1971}, with the distinction that it employs time-varying features and considers only the most recent event to predict the next one, while Hawkes processes include a time-varying increment for each previous event in the history.

\section{\fram: Hawkes for temporal network}\label{sec:approach}
In this section, we show how to apply the previous procedure to
derive a temporal graph generative model and denote this framework \framework.
\begin{figure}[t!]
  \centering
  \includegraphics[width=0.8\textwidth]{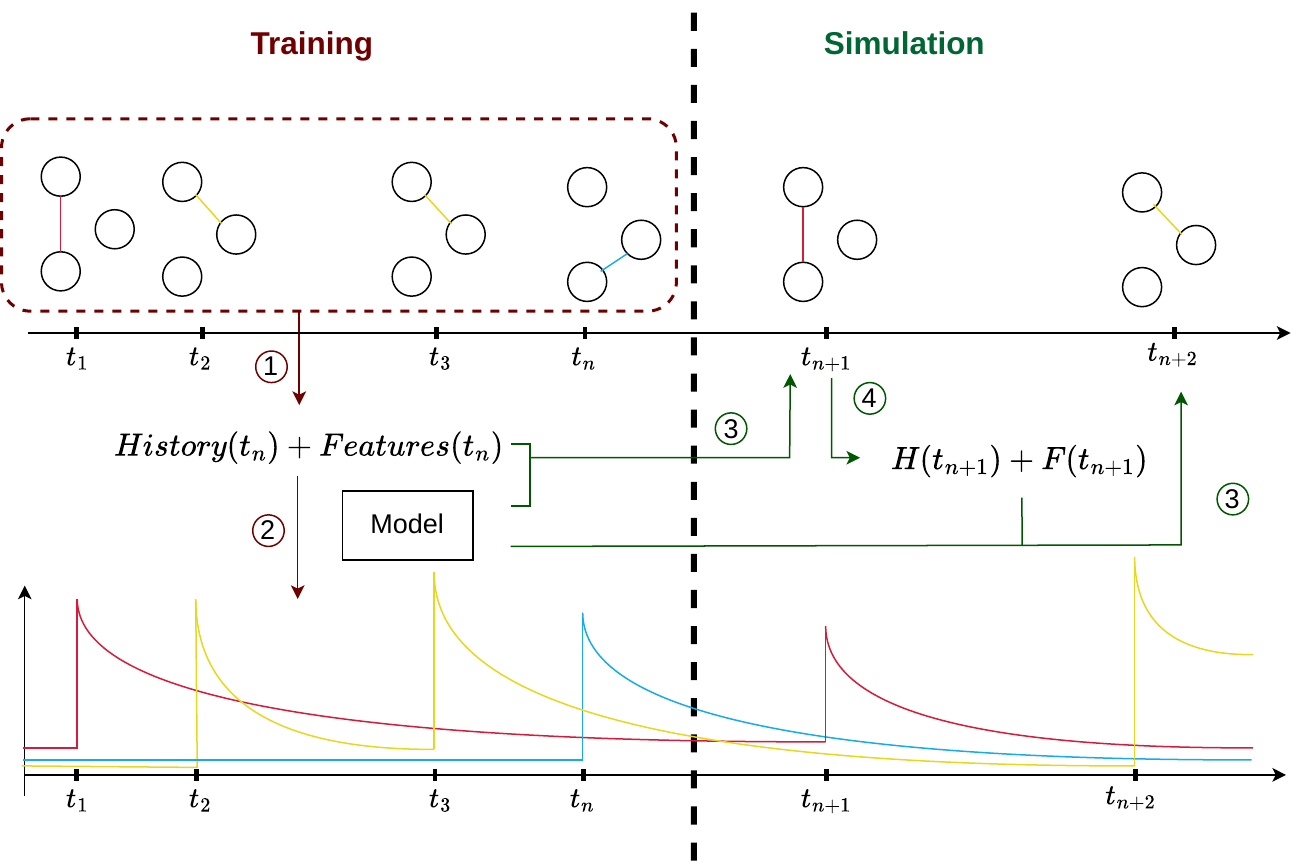}
  \caption{\us
  : \textcolor{red!70!black}{\ding{192}} extraction of the temporal graph's history and hand-crafted creation of features, \textcolor{red!70!black}{\ding{193}} training the model to convert the graph into edge-wise intensity, \textcolor{green!40!black}{\ding{194}} Ogata prediction of the next activated edge and the corresponding time based on the trained model, graph history and features, \textcolor{green!40!black}{\ding{195}} updating the graph's history and features.}
  \label{fig:framework}
\end{figure}
\subsection{Modeling Edge-wise Intensity}
As in previous work \citep{Vu2011,Perry2013e, DuBois2013}, the edges are considered as conditionally independent given the rest of the model. Thus, we start by applying the formulation from Section \ref{markov} at the edge level. This will set the stage for defining the loss as the negative log-likelihood for the entire temporal network. 

In our framework, given a temporal network, for each edge, $e\in \mathcal{E}$ we define the edge-wise conditional intensity function $t\mapsto \lambda^{(e)}(t|\mathcal{H}^{(e)}_t)$, associated to the point process generating the edge history $\mathcal{H}^{(e)}$ :


$$\lambda^{(e)}(t| \mathcal{H}^{(e)}_t) = \mu^{(e)} + \alpha^{(e)} e^{-\beta^{(e)}(t-t^{(e)}_{last})},$$
where $\alpha$ determines how significant the activation of the edge is, $\beta$ is a decay rate for each edge which models how long the interaction lasts, and $t_{last}$ is the time of the last activation of the edge $e$ before $t$ (equivalent to $t_n$ in Equation \ref{eq:hawnet}).
The global intensity of the graph is the sum of these intensities : 
$$ \lambda(t| \mathcal{H}_t) = \sum_{e\in  \mathcal{E}} \mu^{(e)} + \new{\alpha^{(e)}} e^{-\new{\beta^{(e)}}(t-t^{(e)}_{last})} $$
A graph is an object that cannot be studied edge by edge without taking into account their mutual influence, we introduce the notion of time-varying features $F^{(e)}(t) \in \mathbb{R}^{(d)}_+$, which enables us to represent the influence of edges on each other by incorporate it in $\alpha$. Defining these features allows us to describe the edge history $\mathcal{H}^{(e)}_t$ as the set of features and the timestamps of the previous activations of the edge. Finally, the model writes 

\begin{equation}\label{eq:model}
\lambda^{(e)}(t| \mathcal{H}_t) = \mu^{(e)}+ \sigma(g(F^{(e)}(t)) e^{-\beta^{(e)}(t-t_{last})}
\end{equation}
where $g$ is a linear function, $\sigma$ the softplus activation function, and $F(t)$ time-varying features. An example of these features is presented in Section \ref{sec:t_feat}. Finally, $\beta$ is a learned parameter that controls the speed at which an activation loses importance and influence over other edges. Now our goal is to learn the parameters involved in this intensity function, namely the node embeddings, $\beta$, and the parameters of $g$ weighting the features.\\

\noindent\ptitle{Base intensity} The term $\mu^{(e)} $, which defines the base intensity is independent of time. It is calculated as a function of the distance between the embeddings of the two nodes that are involved in the edge, that allows us to capture the static information of the graph. Since our model operates on edges rather than nodes, we derive edge embeddings from the distance between the node embeddings implied in the edge. Without prior knowledge on the temporality of the edge base rate, a simple approach is to define the constant intensity of the edge $e = (u, v)$ as:
\begin{equation}\label{eq:poisson}
    \mu^{(e)} = e^{(2c  - d(z_{u},z_{v}))},
\end{equation}
where for each node $u \in U \cup V $, $z_u \in  R^p$ is an embedding of dimension $p$, $c \in R$ is an offset parameter and $d$ is a distance function.  This model is inspired by the literature about latent space models for graphs. The underlying intuition is that each node $u \in U$ in the network is equipped with a position $z_u$ in a latent social space, and a disk centered around $z_u$ with a radius $c$. $c$ is learned and corresponds to the influence radius of a node in the latent space. So two nodes $u$ and $v$ have an interaction value greater than 1 whenever the intersection of their disks is not empty. In Equation \ref{eq:poisson}, the smaller $d(z_u, z_v)$ is, the higher the base intensity between $u$ and $v$ will be. \\

\textit{Remark}: $\mu$ is highly flexible and can be defined in other ways depending on the aspects we want to incorporate into the model. For example, if we want to add a label to the edges or prioritize certain nodes belonging to a particular community, it is easy to integrate it into $\mu$.


\subsection{Definition of the Loss and Optimization}
Assuming a parametrization of the intensity function above, we propose to find the maximum likelihood estimates of the parameters by optimizing the negative log-likelihood.
Given a point process model with conditional intensity $\lambda(t|\mathcal{H}_t)$, the general form of the likelihood of an observed event history $\left\{t_1,...,t_n\right\}$ on an interval $[0,T]$ is derived in \citet{Daley2010} and given by
\begin{align*}
	p(\left\{t_1,...,t_n\right\})=
	\left(\prod\limits_{i=1}^n \lambda(t_i|\mathcal{H}_t)\right)
	\exp\left(-\int\limits_{0}^T \lambda(s|\mathcal{H}_s)ds\right).
\end{align*}

We use the independence property of the edges conditional on the time-varying features to define the negative log-likelihood of the model:
\begin{definition}
	The negative log-likelihood of our network point process model is given by
	\begin{align}\label{eq:loss}
		L(\mathcal{H})=
		\sum\limits_{e\in \mathcal{E}} \left[\int\limits_{0}^T \lambda^{(e)}(s|\mathcal{H}_s)ds - \sum\limits_{\tau\in\mathcal{H}^{(e)}}
		\log\left(\lambda^{(e)}(\tau|\mathcal{H}_{\tau})\right)
  \right].
	\end{align}
\end{definition}

\newcommand{\Lpos}{L_{(pos)}}
\newcommand{\Lneg}{L_{(neg)}}
Alternatively, this loss function can be re-indexed as a sum indexed by the events, since we have $$L(\mathcal{H})= \Lpos(\mathcal{H}) + \Lneg(\mathcal{H})$$ with:
\begin{itemize}
    \item $\quad \Lpos(\mathcal{H}) = \sum\limits_{(e_m,t_m) \in \mathcal{H}} -\log\left(\lambda^{(e_m)}(t_m|\mathcal{H}_{t_m})\right),$
\item $\quad \Lneg(\mathcal{H}) = \sum\limits_{(e_m,t_m) \in \mathcal{H}} \left( \sum\limits_{e \in \mathcal{E}} \int\limits_{t_{m-1}}^{t_m} \lambda^{(e)}(s|\mathcal{H}_s) ds \right).$
\end{itemize}

To get to this expression, we split the integral on the left-hand side of Equation \ref{eq:loss} over the inter-arrival time interval $\{[t_{m-1}, t_m), m=1,..., M\}$, and we swapped the sum on $w_m$ and the sum over $e$.

\newcommand{\Lnegtilde}{\tilde{L}_{(neg)}}
Although the edge sum $\sum\limits_{e\in \Ecal}$ necessary to compute $\Lneg(\mathcal{H})$ leads to a computational complexity proportional to the square of the number of nodes, the loss function can be approximated by sampling a set of contrasting edge samples $\tilde{\mathcal{E}}_m$ for each event $m$, such that $|\tilde{\mathcal{E}}_m| \ll |\mathcal{E}|$.
We can then substitute the full sum with the normalized mean of their associated survival terms to approximate $L(\mathcal{H}) \approx \Lnegtilde(\mathcal{H}) + \Lpos(\mathcal{H})$ with $$
	\Lnegtilde= \sum\limits_{(e_m,t_m)\in \mathcal{H}}
	\frac{|\mathcal{E}|}{|\tilde{\mathcal{E}}_m|}
	\sum\limits_{e \in \tilde{\mathcal{E}}_m} \int\limits_{t_{m-1}}^{tm} \lambda^{(e)}(s|\mathcal{H}_s)ds).
$$


This approximated loss function has a number of terms proportional to the number of events in the network history.
It can be further approximated by computing it on batches, formed by taking slices of the history.

\subsection{Temporal features}{\label{sec:t_feat}
Hereafter, we provide examples of handcrafted features composing $F^{(e)}$. It is worth noting that these features can be learned in practice \citep{WangCLL021}, but we have chosen not to explore this direction in this work, and to leave it as an open question (see Conclusion).
As is common in network analysis, the choice of features is highly dependent on the nature of the network considered, and the type of analysis being conducted.
In a variety of previous work, some time-varying edge features have been introduced \citep{Butts2008,Vu2011,DuBois2013}.
We propose to exploit three of them in this work, which consists of an exponentially down-weighted version of the volume ($\mathbf{vol}$), the preferential attachment ($\mathbf{pa}$), and common neighbors ($\mathbf{cn}$).

Here we denote by $\gamma$ a decay hyper-parameter controlling the rate at which the feature contributions from previous events become obsolete. In addition, $CN(u,t)$ represents the neighborhood for a given node $u$ at time $t$ (i.e., the set of nodes that had an interaction with node $u$ before $t$).Then these features are written as follows

\begin{align*}
    \mathbf{vol^{(e)}(t)}&=\sum_{
			      \substack{t^{(e)}_i < t}
		      }\exp(-\gamma(t-t_i^{(e)})), \quad \mathbf{cn^{(e)}(t)}=\sum\limits_{w \in CN(e,t)} \exp(-\gamma(t-t_{last}^{(e)}(w)))\\
         \text{and} & \quad \mathbf{pa^{(e)}(t)} = deg_u * deg_v, \quad\text{with}\quad deg_u(t) = \sum_{n_u \in CN(u,t)} \exp(-\gamma(t-t_{last}^{n_u})).
\end{align*}

$\mathbf{vol^{(e)}(t)}$ measures the volume of recent interactions, where $t_i^{(e)}$ denotes the time of the previous activations of the edge $e$. We also have that $CN(e,t) = CN(u,t)\cap CN(v,t)$ is the set of neighbors of the edge $e = (u,v)$  at time $t$, the set of nodes that have created an edge before time $t$ with the two nodes $u$ and $v$. Moreover, $t_{last}^{(e)}(w)$ is the time of the last interaction between $u$ or $v$, and the common neighbors $w$.

\begin{figure}[!h]
    \centering
    \begin{subfigure}[b]{0.44\textwidth}
        \begin{tikzpicture}[scale=0.53] 
  \node[draw=red, circle, scale=0.8] (A) at (2,3) {A};
  \node[draw =red, fill=none, circle, scale=0.8] (B) at (5,0) {B};
  \node[draw, circle, scale=0.8] (C) at (7,3) {C};
  \node[draw, circle, scale=0.8] (D) at (0,0) {D};
  \node[draw, circle, scale=0.8] (E) at (9,0) {E};
  
  \tikzset{EdgeStyle/.append}
  \Edge[label = $e_4$](A)(C);
  \Edge[label = $e_3$](E)(C);
  \Edge[label = $e_2$](B)(D);
  \Edge[label = $e_1$](B)(E);
  \Edge[label = $e_5$](A)(B);
  \tikzset{EdgeStyle/.style = {red}}

  \node[above=0.1cm,  font=\tiny,text=blue] at (4.5,3) {$t_{e_{4}} = (2,8)$};
  \node[above=0.2cm,  font=\tiny,text=blue] at (9,1) {$t_{e_{3}} = 6$};   
  \node[above=0.05cm,  font=\tiny,text=blue] at (7,-1) {$t_{e_{1}} = (5,10)$};   
  \node[above=0.05cm, font=\tiny,text=blue] at (2.5,-1) {$t_{e_{2}} = 4$}; 
  \node[above=0.2cm, left=-0.5cm, font=\tiny,text=blue] at (2,1) {$t_{e_{5}} = (1,3,7)$}; 
\end{tikzpicture}
    \end{subfigure}
    \begin{subfigure}[b]{0.25\textwidth}
    \includegraphics[width=\textwidth]{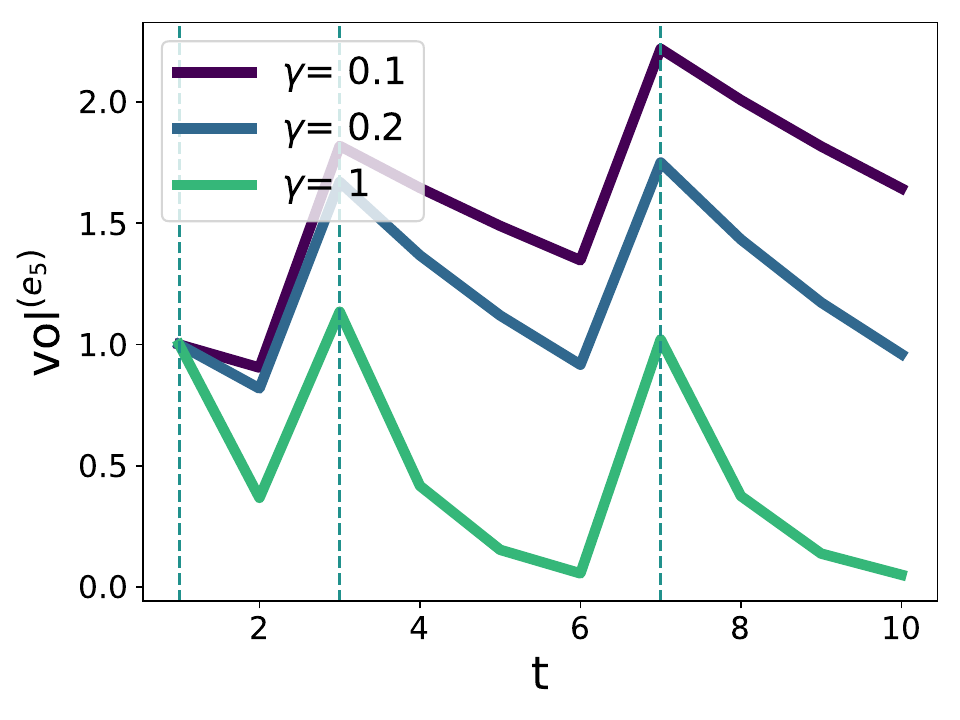}
    \end{subfigure}
     \begin{subfigure}[b]{0.25\textwidth}
    \includegraphics[width=\textwidth]{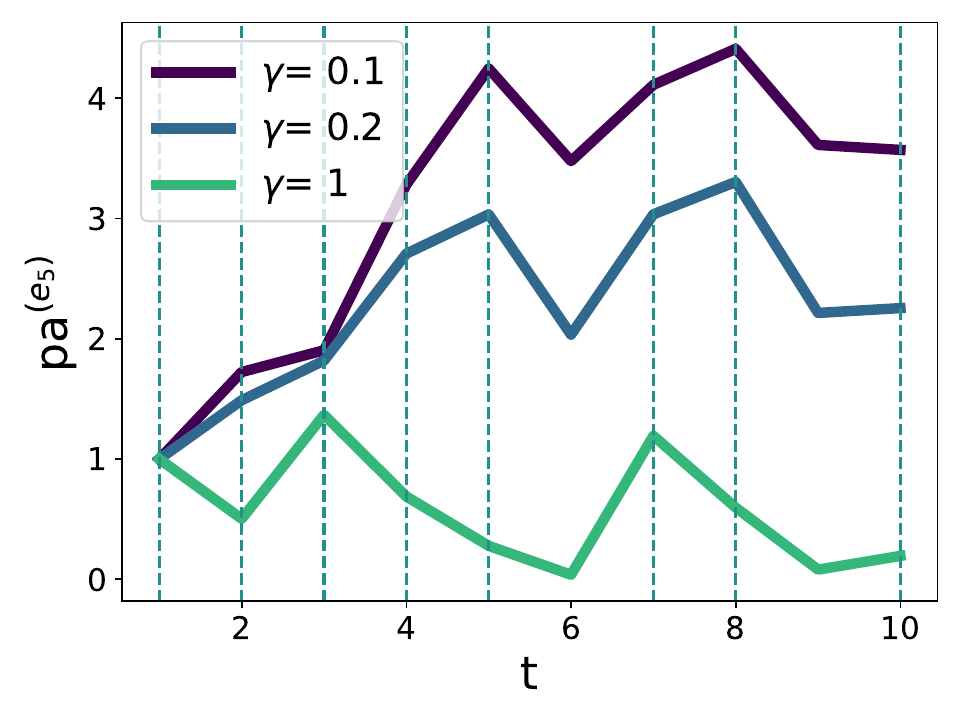}
    \end{subfigure}
    \caption{(left) Example of a toy graph and temporal features computed on $e_5$, namely (middle): the volume, $\mathbf{\text{\textbf{vol}}^{(e_5)}}$ and (right) the preferential attachment, $\mathbf{\text{\textbf{pa}}^{(e_5)}}$ as functions of time $t=1, \cdots, 10$. Vertical lines represent the time of activation of edges involved in the computation of each feature.}
    \label{fig:toy-graph}
\end{figure}

\ptitleskip{Feature computation example}
Figure \ref{fig:toy-graph} shows a toy graph that illustrates the differences between the features described above. The toy graph is composed of five nodes: $\mathcal{G}=\{A, B, C, D, E\}$. Edges are denoted by $e$ and for each edge we record their time of activation. For instance, if the graph corresponds to email communication between individuals within a company, then A and B exchange three emails in the observed period, at times 1, 3, and 7.

Focusing on $e_5$, we can compute the features for this particular edge along the time, so from $t=1$ to $t=10$. We vary the value of $\gamma$ to better understand its impact. In Figure \ref{fig:toy-graph}, we present the volume and the preferential attachment features computed for $e_5$.
The volume only depends on the edge itself, and not on any other edges of the graph. As such, we can see that it reaches peaks when $e_5$ gets activated, i.e., at times 1, 3 and 7. Between these activation times, the volume decreases at a rate that is controlled by the value of $\gamma$. Preferential attachment on the other hand is also impacted by the activations of edges that belong to either the direct neighborhood of $A$ or $B$. For this reason, we observe peaks corresponding to the activation of $e_5$ but also to the activations of $e1, e2$, and $e_4$.

\textit{Remark}: $\gamma$ impacts both the speed at which the values of the temporal features decreased toward $0$ and the scale of the features. 

\section{\us: From Intensity Functions to Graph Simulation}\label{sec:sim}
We are now ready to use the intensity functions defined in the previous section to simulate dynamic networks. The main idea is to treat the graph as a composition of intensity functions, enabling us to use rejection sampling to generate observations. By modeling the intensity of interactions as a function of both the network structure and temporal dynamics, we can efficiently simulate realistic event sequences that mimic the behavior of the original temporal graph.

\subsection{Simulation}

Inspired by the thinning algorithm presented by \citet{Ogata1981}, we propose the following simulation algorithm.
For every edge $e$, we remind that $t \mapsto\lambda^{(e)}(t|\mathcal{H}_t)$ denotes the conditional intensity of the point process of edge $e$. Similarly, we define $\lambda$, the total conditional intensity function, that generates the sequence of event times in the network:

$$t \mapsto \lambda(t|\mathcal{H}_t) = \sum\limits_{e \in \mathcal{E}} \lambda^{(e)}(t|\mathcal{H}_t).$$

The full pseudocode is given in the Appendix, Algorithm \ref{alg:ogata}. We describe the main steps hereafter.\\

\noindent\ptitle{Step 1 -- Generation of a new time} 
Starting from an initial time $t_n$, we fix the final time at which the simulation must stop and a maximum number of iterations $N$. At each iteration, it generates a random number that will be used to determine the time until the next event $\Delta_t$. This time is based on the minimum of an exponential process depending on edge intensities $\lambda^{(e)}(t)$ and a variable $u \sim \mathcal{U}(0,1)$ :
$$\Delta_t = \min\limits_{e}\left(- \frac{\ln(u^{(e)})}{\lambda^{(e)}(t)}\right).
 $$
  Secondly, we compute a local upper bound, $\lambda^{(*)}$ of the total intensity on the right neighborhood of the current time at each iteration. This upper bound allows us to validate the next event time at time $t_n + \Delta_t$ across the network, in a rejection-sampling manner.

\textit{Remark}: We choose to stop the simulation at a specific time instead of basing it on the number of activated edges. Although this second option is feasible, it requires knowing in advance the number of activated edges, which is typically not realistic.\\



\noindent\ptitle{Step 2 -- Probability computation} Once a new event time is generated, we need to assign it to an edge in the network. To do so we simulate a set of activatable edges $\mathcal{E}_{\text{active}}$ at this new time using rejection sampling based on their respective intensities. We then compute the activation probability for each edge at this time. This probability is proportional to their intensity value and given by
\begin{equation}\label{eq:p}
  p^{(e)}(t) = \frac{\lambda^{(e)}(t | \mathcal{H}_{t})}{\sum_{e' \in \mathcal{E}_{\text{active}}}\lambda^{(e')}(t | \mathcal{H}_{t})}.  
\end{equation}


\noindent\ptitle{Step 3 -- Activation of a new edge} We sample a top-$k$ of edge candidates having the highest likelihood to be activated, such that $e_{\text{new}} \sim \text{Multinomial}(1, \{ p_e | e \in \epsilon \})$.
This strategy is motivated by the risk of always selecting the same edge resulting in some infinite loop. By picking $k$ large enough we are forcing a more distributed activation pattern over the network.
\\

\noindent\ptitle{Step 4: Updating} Finally, the activated edge is selected uniformly at random among the $k$ sampled ones.
The event is added to the history and updated as the activation occurs. \\

\noindent\ptitle{On the importance of time normalization} The Ogata algorithm is highly sensitive to the time range over which events are observed. This sensitivity arises from the algorithm's method of calculating $\Delta _t $ the time interval until the next potential event occurrence. On one hand, if the time range is too large, $\Delta _t $ becomes excessively small, leading to an explosion in the number of generated events. On the other hand, if the time range is too small, the number of generated events becomes insufficient. 

\subsection{Limitations of \us}
The main limit of \us~ is that the used features focus on the neighborhood of the edge and its historical context; however, it lacks global features that could provide additional insights into the overall trends of the graph. For instance, for some applications, factors such as whether it is a weekend or a weekday could significantly impact the overall activity on the graph. We also note that with each new event accepted into the network, the variables used to calculate time-varying features need to be updated. For example, $\mathbf{vol^{(e)}(t)}$ needs access to all timestamps of previous time events for a given edge. The complexity of this update operation depends on the feature under consideration but can also prevent the algorithm from scaling to larger graphs. 

One way to address both issues could be to use a more sophisticated approach to learn the temporal features. We discuss this avenue of research in Section \ref{sec:conclusion}.\\

\noindent\ptitle{Summary} We summarize the full pipeline (model training and simulation) in Figure \ref{fig:framework}. The pipeline can be decomposed into the following steps. We start by training our model, i.e. we optimize Equation \ref{eq:loss} based on the observed events and the handcrafted features. This model allows us to convert the graph into edge-wise intensity functions. Then, based on edge-wise intensity functions obtained, we predict the evolution of the graph following \us. 

\section{Experiments}\label{sec:xp}
Now, we present the evaluation framework that we designed to assess the capability of \us~to capture and then replicate the evolution of a network. We also present numerical results obtained within this framework. The code for conducting all experiments is available online$^{\ref{ftn:code}}$.
\subsection{Experimental Protocol}
To evaluate the ability of our approach to capture the dynamic of the network we use both real-world networks and synthetic graphs.\\

\noindent\ptitle{Synthetic Networks}
For synthetic graphs, we create four distinct scenarios inspired by different models. The configuration model \citep{newman2003}, based on a given degree sequence allows us to model influence effects within a network (i.e., the more influence a person has, the higher the degree of the associated node). The second uses stream graph \citep{LatapyFZ19} combined with the Erd\H{o}s-R\'enyi Model. The two final scenarios leverage Stochastic Block Model to simulate the creation of community structures within the network. The scenario referred to as SBM-a is a temporal graph composed of two strong communities and the SBM-b extends this by modeling the dissolution of one of the communities, followed by the emergence of a new one. Additional details regarding the underlying generative process of each of these graphs are provided in Section \ref{ref:SGG}. 


\begin{figure}[!h]
  \centering
  \includegraphics[width=1\textwidth]{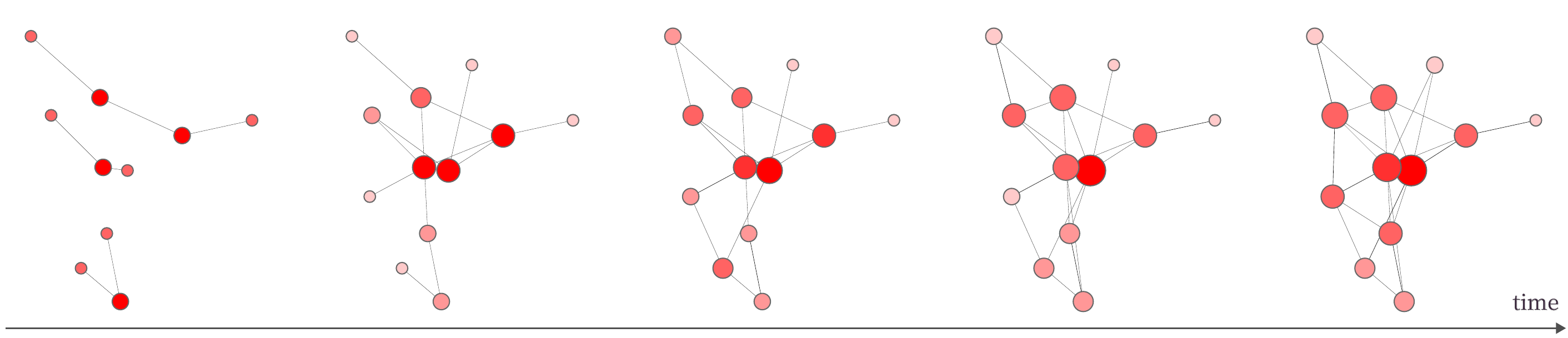}\\
  \includegraphics[width=\textwidth]{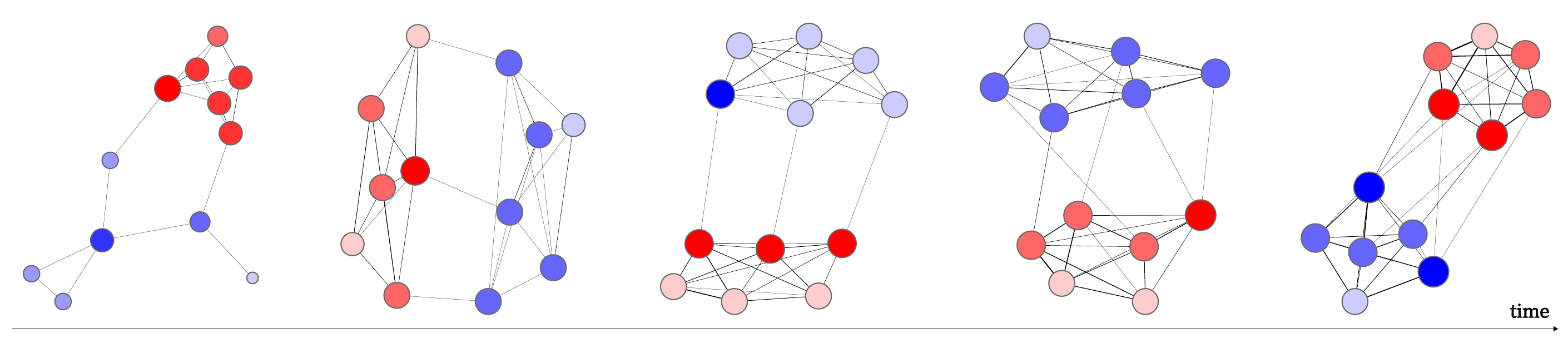}
  \caption{Temporal snapshot of synthetic dynamic graphs generated based on the (top row) stream graph and (bottom row) SBM. The darker the node color, the higher its degree. For SBM blue and red nodes represent the community.}
  \label{fig:config}
\end{figure}

\noindent\ptitle{Real Networks}
We use three real-world networks : 
the Enron Dataset\footnote{\url{https://www.cis.jhu.edu/~parky/Enron/}} that represents direct messages sent between employees of the company Enron between 1999 and 2003, Reality Mining \citep{10.1007/s00779-005-0046-3} which is communication recorded continuously over 50 week and Militarized Interstate Disputes
(MID) \citep{doi:10.1177/0738894221995743} which records all instances of when one state threatened, displayed, or used force against another between 1816 and 2014 (more details in Appendix \ref{ref:NETprop}}). \\

\noindent\ptitle{Baselines} We compare our model with two other continuous-time network models based on generative temporal point processes: the Latent Space Hawkes (\textsc{LSH}) model \citep{huangMutuallyExcitingLatent2022} and the Hawkes Dual Latent Space (\textsc{DLS}) model \citep{yang2017decoupling}. \textsc{LSH} utilizes a latent space representation for nodes and models interactions through mutually exciting Hawkes processes with baseline intensities influenced by node distances and sender-receiver-specific effects. \textsc{DLS} combines homophily and reciprocal latent spaces to account for how users initiate communication and exchange dynamics, utilizing a mixture of exponential and periodic kernels.\\

\noindent\ptitle{General setup} We simulate synthetic graphs with 100 nodes and a varying number of interactions depending on the model used. For each type of synthetic graph (configuration model, SBM, and stream graph) we generate 10 initial graphs then on each generated graph, we run our approach 10 times to evaluate the robustness of the \us~ simulation. For both real-world and synthetic graphs, we perform the train-test split based on time, where the first 70\% of interactions are used for training and the remaining ones for testing. Statistics about the dataset are provided in the Appendix, Table \ref{tab:dataset} and Figure \ref{fig:config} showcase examples of snapshots of the dynamic graphs generated by these processes.

\subsubsection{Evaluation metrics}
A significant challenge for our work is the evaluation of \us.  Our objective is to comprehensively assess the ability of \us\ to capture the static component of the graph (i.e., determine the pair of nodes that will connect) and the dynamic part (i.e., in which particular order and when these connections will happen).\\

\noindent\ptitle{Static metrics} Following \cite{huangMutuallyExcitingLatent2022}, we report the clustering coefficients, the number of events (number of activations) and the average degree of the simulated graphs. We also report the accuracy that is the ratio of the number of correctly predicted edges to the total number of edges simulated. This metric can be seen as a form of true positive rate for edge activation.

\begin{figure}[!h]
    \centering
    \begin{subfigure}[b]{0.32\textwidth}
        \includegraphics[width=\textwidth]{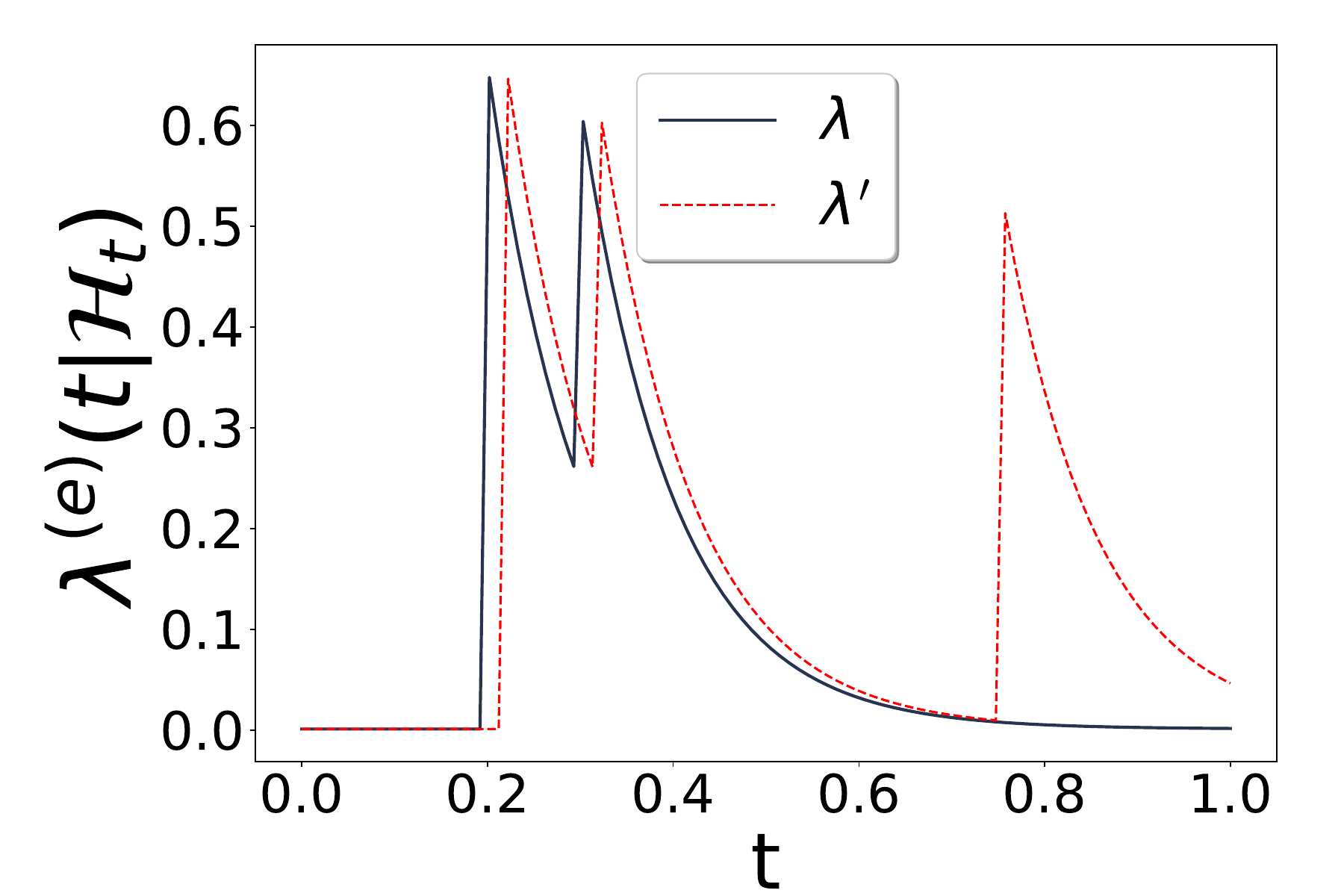}
        \caption{$DTW = 0.91$}\label{fig:dtw_a}
    \end{subfigure}
    \hfill
    \begin{subfigure}[b]{0.32\textwidth}
        \includegraphics[width=\textwidth]{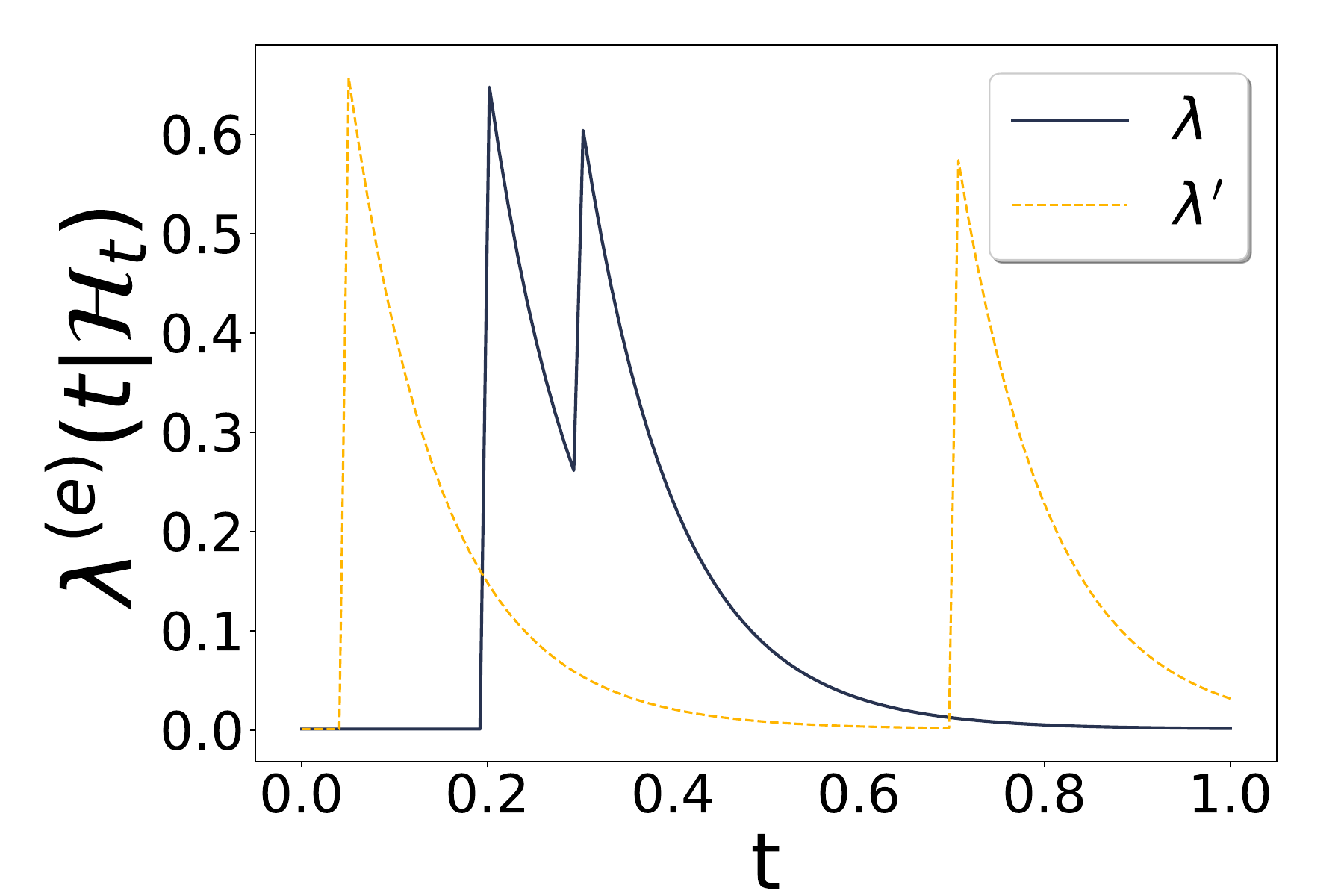}
        \caption{$DTW = 1.17$}\label{fig:dtw_b}
    \end{subfigure}
    \hfill
    \begin{subfigure}[b]{0.32\textwidth}
        \includegraphics[width=\textwidth]{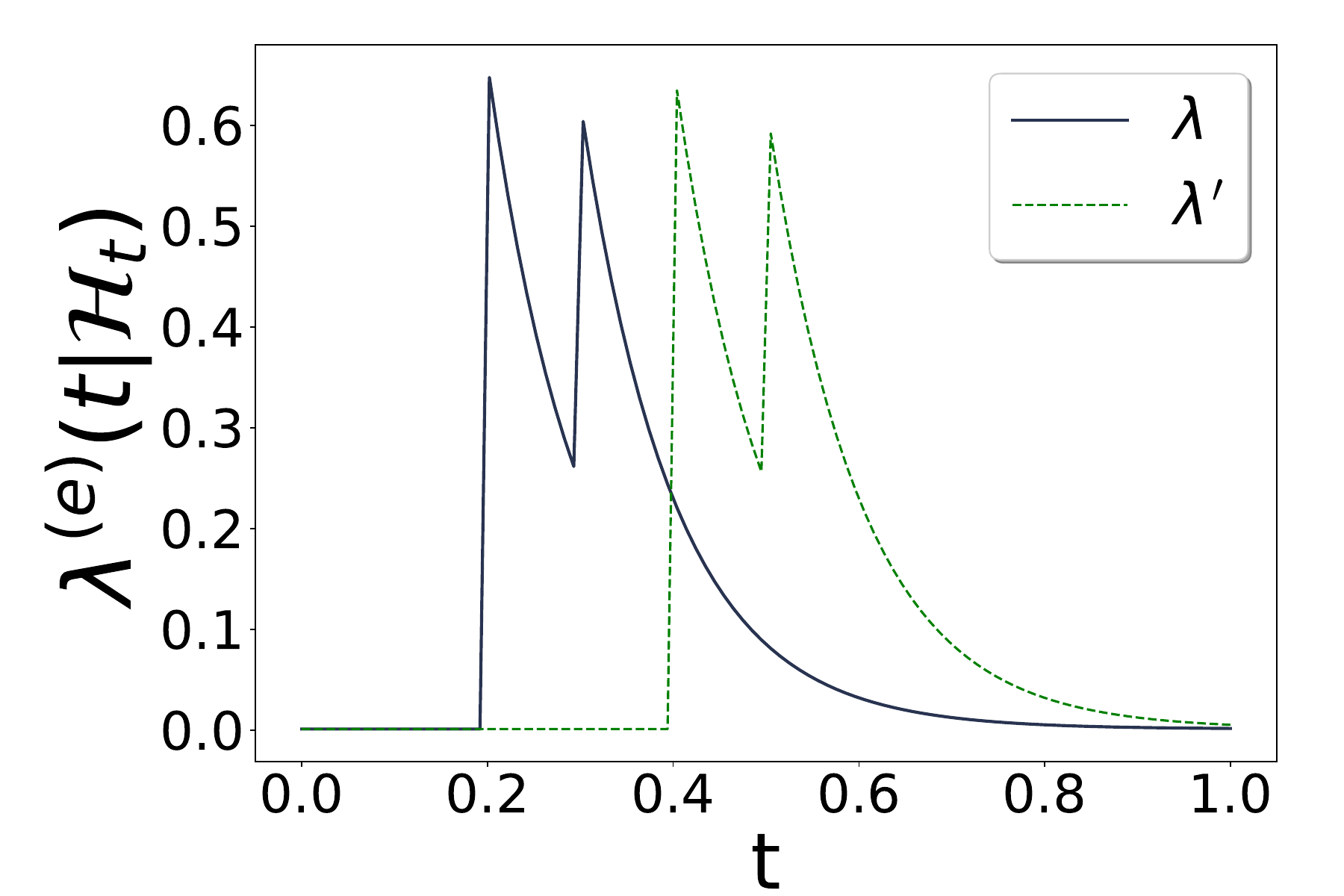}
        \caption{$DTW = 0.04$}
    \end{subfigure}
    
    \caption{Examples of DTW values between edge intensity functions; the ground truth $\lambda$ is the same across all examples and $\lambda'$ represents (a) 3 activations instead of two; (b) the same number of activations but not in the same time theme and (c) the same number of activations but slightly shifted.
    }
    \label{fig:visu-dtw}
\end{figure}

\noindent\ptitle{Dynamic metric} For the dynamic component, our goal is to evaluate the temporal coherence of our simulation. We proceed as follows: once our model is trained we extract \emph{the intensity function} computed for each edge both on the original graph and on the graph obtained with our fixed-parameter model. Formally, given $\lambda^{(e)}_\mathcal{G}\in \mathbb{R}^T$ the edge-wise intensity distribution over time for an edge $e$ in a given graph $\mathcal{G}$, we can evaluate at a global level by computing the graph intensity as $\lambda_{\mathcal{G}}=\sum_{e}\lambda^{(e)}_\mathcal{G}$. The distance between the global intensity of two graphs $\mathcal{G}$ and $\mathcal{G'}$ is then defined by
$$\text{DTW}^{\text{global}} = DTW(\lambda_{\mathcal{G}}, \lambda_{\mathcal{G'}}).$$
where DTW corresponds to the Dynamic Time Warping (DTW) distance \citep{vintsyuk1968,saoke78}, $\mathcal{G}$ is the original graph and $\mathcal{G'}$ is the graph obtained with ~\us. The motivation behind this proposed metric is to treat the intensity distributions over time as time sequences.

Then, we need an alignment-based metric, that relies on a temporal alignment of the sequences to assess their similarity. In this context, the popular Dynamic Time Warping (DTW) metric emerges as a natural candidate, as it can effectively accommodate minor shifts in the activation timing. An illustration of the value of the DTW between edge intensity functions in three different scenarios is provided in Figure \ref{fig:visu-dtw}. We can see that the DTW is appropriate to assess temporal coherence as it penalizes more cases where the number of predicted activations is wrong for a given edge (Figure \ref{fig:dtw_a}) and when the waiting times between activations are far from the ground truth (Figure \ref{fig:dtw_b}). 
A DTW of $0$ indicates perfect similarity.


\subsection{Results}

\subsubsection{Generation Accuracy}
Table \ref{tab:results_simu} shows the results for all metrics on the synthetic networks (for more information, in Appendix \ref{app:clustering_coef}, we present the degree centralities of the nodes). Note that we do not present the results obtained with \textsc{LSH} because, despite our best efforts, we were unable to find a combination of hyperparameters that would give reasonable results.

We find that our model tends to overestimate the number of predicted edges, whereas DLS underestimates this count. In terms of graph clustering---which reflects the graph's structural organization---\textsc{\us} outperforms DLS. Regarding the temporal aspect, as measured by global DTW, both models exhibit comparable performance. Finally, when it comes to accuracy, DLS outperforms our model. \\

\begin{table}[!h]
    \small
    \centering
    \resizebox{\textwidth}{!}{
        \begin{tabular}{lccccc}
            \toprule
            Network & Approach & \# events & CC. &DTW$^{\text{glob.}}$&Accuracy \\
            \midrule
            \multirow{3}{*}{Configuration} & Ground truth & $700 \pm 20$ & $0.1 \pm 0.02$ &$0$&$100$\\
            & \textsc{DLS} & $529 \pm 41$ & $0.17 \pm 0.02$ &$129 \pm 16$ & \textbf{9 $\pm$ 1} \\
            & \us & \textbf{832 $\pm$ 165 } &  \textbf{ 0.06 $\pm$ 0.02} & \textbf{100 $\pm$ 26} & $ 5 \pm 1$ \\
            \midrule
            \multirow{3}{*}{Stream} & Ground truth &$577 \pm 22$& $0.06 \pm 0.01$&$0$ &$100$\\
            & \textsc{DLS} & \textbf{648 $\pm$ 84 } &$0.18 \pm 0.01$ &\textbf{ 81 $\pm$ 6 } &\textbf{ 23 $\pm$ 6} \\
            & \us & $ 1586 \pm 258$ & \textbf{0.07 $\pm$ 0.02} &$ 229\pm 60$ & $ 15 \pm2 $ \\
            \midrule
            \multirow{3}{*}{SBM-a} & Ground truth &$489 \pm 7$& $0.17 \pm 0.01$&$0$&$100$\\
            & \textsc{DLS} & \textbf{375 $\pm$ 17} & $0.13 \pm 0.01 $& \textbf{ 103 $\pm$ 12} & \textbf{16 $\pm$ 1}\\
            & \us & $ 850 \pm  181$ & $ 0.21 \pm  0.05$& $ 125 \pm 37 $& $ 9\pm 1 $ \\
            \midrule
            \multirow{3}{*}{SBM-b} & Ground truth &$698 \pm 8$ & $0.25 \pm 0.02$& $0$&$100$\\
            & \textsc{DLS} &$ 499 \pm 15$& $0.15 \pm 0.01 $& $152 \pm  7$ & \textbf{15 $\pm$ 1}\\
            & \us & \textbf{ 865 $\pm$  200} & \textbf{0.20$ \pm $0.08 }&  \textbf{115 $\pm$ 19 }& $7 \pm 2 $ \\
            \bottomrule
        \end{tabular}
    }
    \caption{Mean $\pm$ std for all metrics on the synthetic networks. {CC} is the clustering coefficient and Accuracy is the ratio of the number of correctly predicted edges to the total number of edges simulated.}
    \label{tab:results_simu}
\end{table}
For the real-world networks discussed below, the authors provided the set of optimal hyperparameters values, so for these we do report the results of \textsc{LSH}.
We present the results obtained on the three real-world networks in Table \ref{tab:results}. First, we observe that \textsc{LSH} significantly overestimates the number of events on two networks, with discrepancies reaching a factor of 4 on Reality and exceeding a factor of 10 on MID. This issue of instability in estimating the number of events was highlighted by \cite{huangMutuallyExcitingLatent2022} for both \textsc{LSH} and \textsc{DLS}.
Interestingly, while \cite{soliman2022multivariatecommunityhawkesmodel} also reported that \textsc{DLS} is unstable and unsuitable for generating novel networks, our independent execution of their code did not confirm this limitation.

\begin{table}[!h]
    \small
    \centering
    \resizebox{\textwidth}{!}{
        \begin{tabular}{lcccccc}
            \toprule
            Network & Approach & \# events & CC &DTW$^{\text{glob.}}$&Accuracy \\
            \midrule
            \multirow{4}{*}{Enron} & Ground truth & $1133$ & $0.23$ & $0$ &$100$&\\
            & \textsc{LSH} & \textbf{1044 $\pm$ 893} &  \underline{$0.3 \pm 0.01$ } &  \underline{$309 \pm 162 $} & $0.1 \pm 0$ & \\
            & \textsc{DLS} & $584 \pm 21 $ & \textbf{0.16 $\pm$ 0.02} & \textbf{247 $\pm$ 16  } & \textbf{33 $\pm $2 } & \\
            & \us & \underline{$1467 \pm 64$} & $0.12 \pm 0.02$ & $399\pm 31$ &  \underline{$9 \pm 1$} & \\
            \midrule
            \multirow{4}{*}{MID} & Ground truth &$1272$ & $0.13$& $ 0$& $100$&\\
            & \textsc{LSH} &$15532 \pm 2253$ & $0.46 \pm 0.03$ & $1220 \pm 195$ & $0$ &\\
            & \textsc{DLS} & \underline{$698 \pm 35$}&\textbf{0.18 $\pm$ 0.03}& \underline{$ 223 \pm 24$ }& \textbf{45 $\pm$ 1}&\\
            & \us &\textbf{1074  $\pm$ 37} & \underline{$0.29 \pm 0.02$ }& \textbf{155 $\pm$ 12} &  \underline{$20 \pm 1$ }&\\
            \midrule
            \multirow{4}{*}{Reality} & Ground truth &$540$&$0.2$&0&100&\\
            & \textsc{LSH} &$888 \pm 723$ &\textbf{0.2 $\pm$ 0.05}& $98 \pm 86 $&$ 0 \pm 1 $ \\
            & \textsc{DLS} &\textbf{568 $\pm$ 27}& \underline{$ 0.19 \pm 0.03$}&\textbf{ 64 $\pm$ 4} & \textbf{60 $\pm$ 1} &\\
            & \us & \underline{$651 \pm 25 $}& $0.25 \pm 0.04 $& \underline{ $ 82\pm  8 $}& \underline{ $29 \pm 2$}&\\
            \bottomrule
        \end{tabular}
    }
    \caption{$\text{Mean}\pm {\text{std}}$ for all metrics on all datasets over 10 runs. {CC} is the clustering coefficient and Accuracy is the ratio of the number of correctly predicted edges to the total number of edges simulated. The best results are highlighted in bold and the second are underlined.}
    \label{tab:results}
\end{table}

\begin{figure}[!h]
    \centering
    \begin{subfigure}[b]{0.49\textwidth}
        \includegraphics[width=\textwidth]{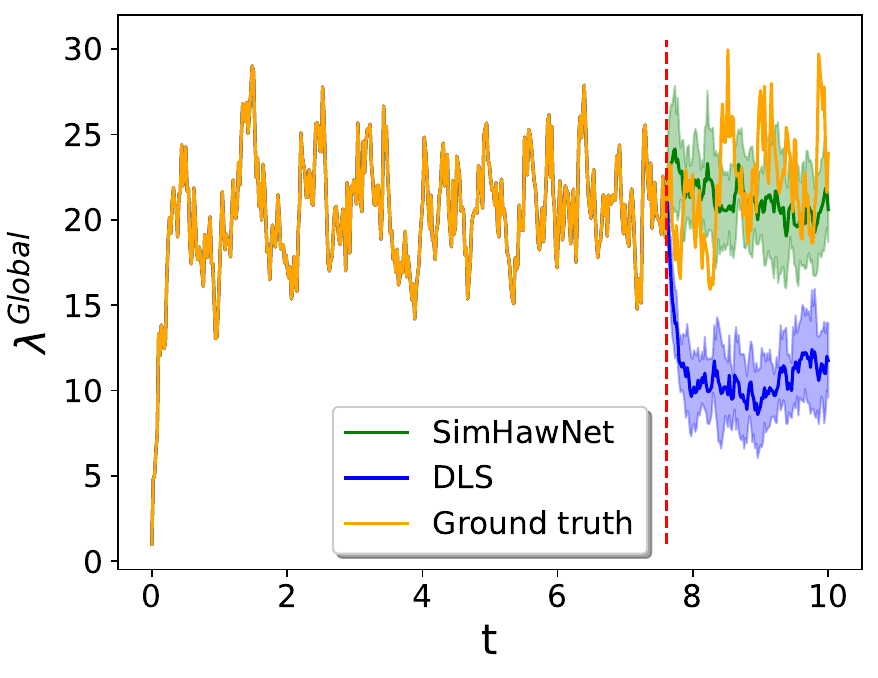}
        \caption{Config}
    \end{subfigure}
    \begin{subfigure}[b]{0.49\textwidth}
        \includegraphics[width=\textwidth]{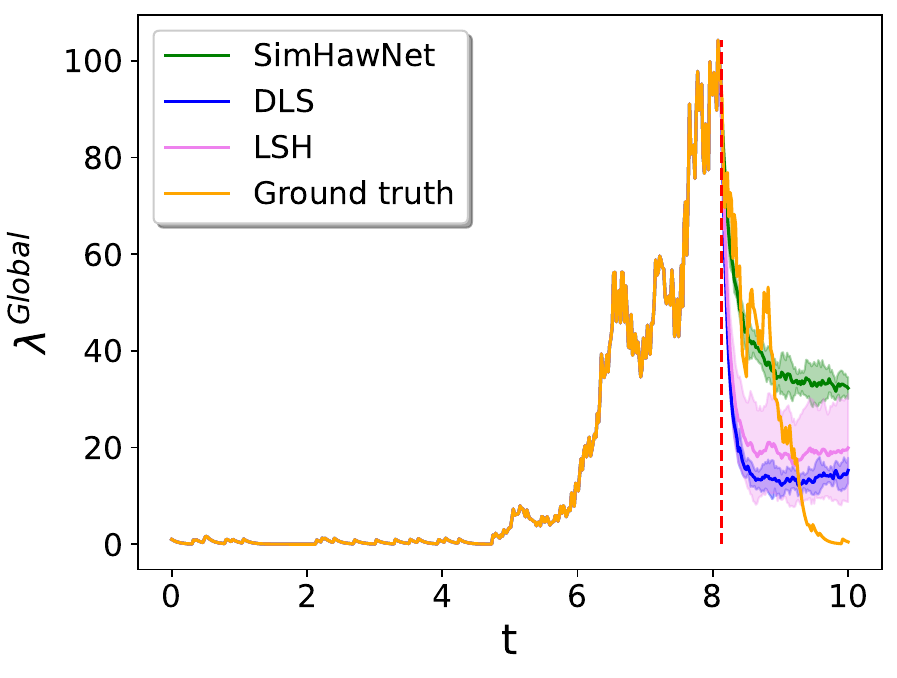}
        \caption{Enron}
    \end{subfigure}
    \caption{Global intensity function obtained. The red line marks the beginning of the simulation.}
    \label{fig:intensity}
\end{figure}

Figure \ref{fig:intensity} presents the intensities resulting from the simulation of 10 graphs using the configuration model dataset for both DLS and \textsc{\us}. On the right, the intensities for the Enron dataset are shown, also including LSH. Additional intensity plots can be found in Appendix \ref{app:intensities}. We observe that our model performs remarkably well across most datasets. Even for MID and Stream, it successfully predicts the peaks; however, it struggles to capture the subsequent declines.

\subsubsection{Parameters analysis}\label{app:param}
Now, we propose to take a closer look at the parameters learned by our model, i.e., the node embeddings learned in the base rate, the weights of handcrafted features (see Appendix \ref{app:param}), and the decay rates $\beta$.

\smallskip

\begin{figure}[!h]
    \centering
    \makebox[\textwidth][c]{%
        \includegraphics[width=1.\textwidth]{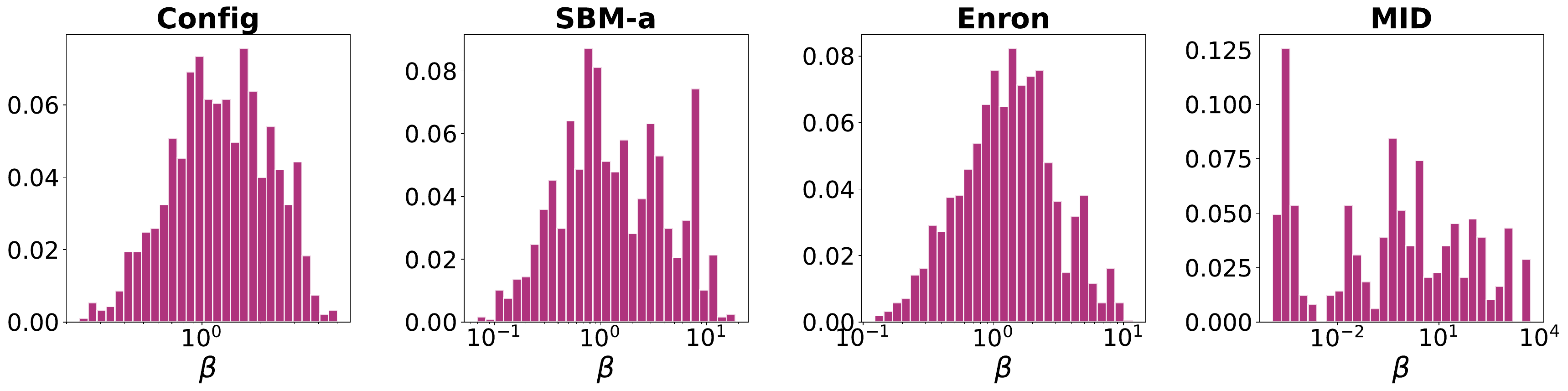}
    }
    \caption{Plot the percentage of the decay ($\beta$) values across the edges for the model on different learned datasets, with a logarithmic scale on the x-axis for $\beta$.}
    \label{fig:beta_distribution}
\end{figure}

\noindent\textbf{Analysis of $\beta$. }Figure \ref{fig:beta_distribution} shows the distribution of the $\beta$ parameters learned by the model. We recall that $\beta$ quantifies how the activation of an edge influences the timing of subsequent activations of the same edge. In other words, if $\beta$ is high, the activation of an edge strongly increases the likelihood of another activation occurring soon afterward on the same edge. In contrast, if $\beta$ is low, the activation of an edge increases the likelihood of another activation occurring over a longer time interval.

For the dataset configurations SBM-a and Enron (as well as SBM-b and Stream, see Appendix \ref{app:param}),we observe that the proportion of values below 1 is almost equal to the proportion of values above 1, excluding extreme cases. This indicates that the model effectively captures variations in the temporal influence of edges. This behavior is consistent with the synthetic data, where extreme edge behaviors—such as frequent repeated activations or isolated activations of an outlier edge—are not present.
In the case of MID (and Reality; see Appendix \ref{app:param}), we observe more extreme values (i.e., closer to $0$ or greater than $10$), suggesting that the model distinguishes more sharply between the impacts of different edge activations. Some activations have very short-term or negligible effects (values close to $10$), while others exhibit influence over extended periods or even throughout the simulation (values close to $0$). This is particularly pronounced in the MID dataset, where values reach as high as $10^3$, indicating that the model interprets certain activations as singular, one-off events, while others have sustained, long-term effects. This differentiation aligns with historical observations: some conflicts are brief and isolated, while others, such as the Cold War, span much longer durations.

\begin{figure}[!h]
    \begin{subfigure}[b]{0.35\textwidth}
        \centering
        \begin{subfigure}[b]{\textwidth} 
            \centering
            \includegraphics[width=0.8\textwidth]{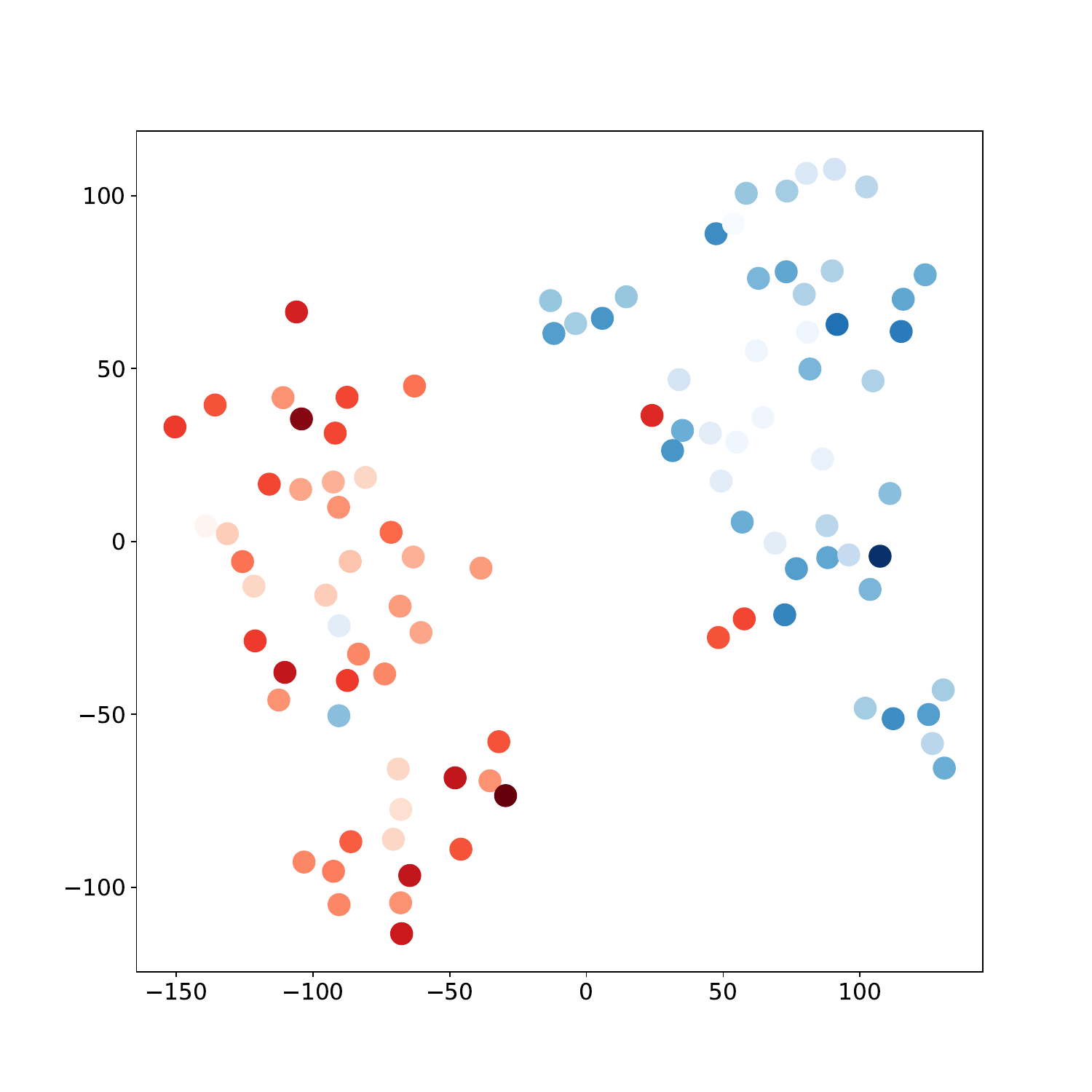}
        \end{subfigure}
        \begin{subfigure}[b]{0.8\textwidth} 
            \centering
            \includegraphics[width=\textwidth]{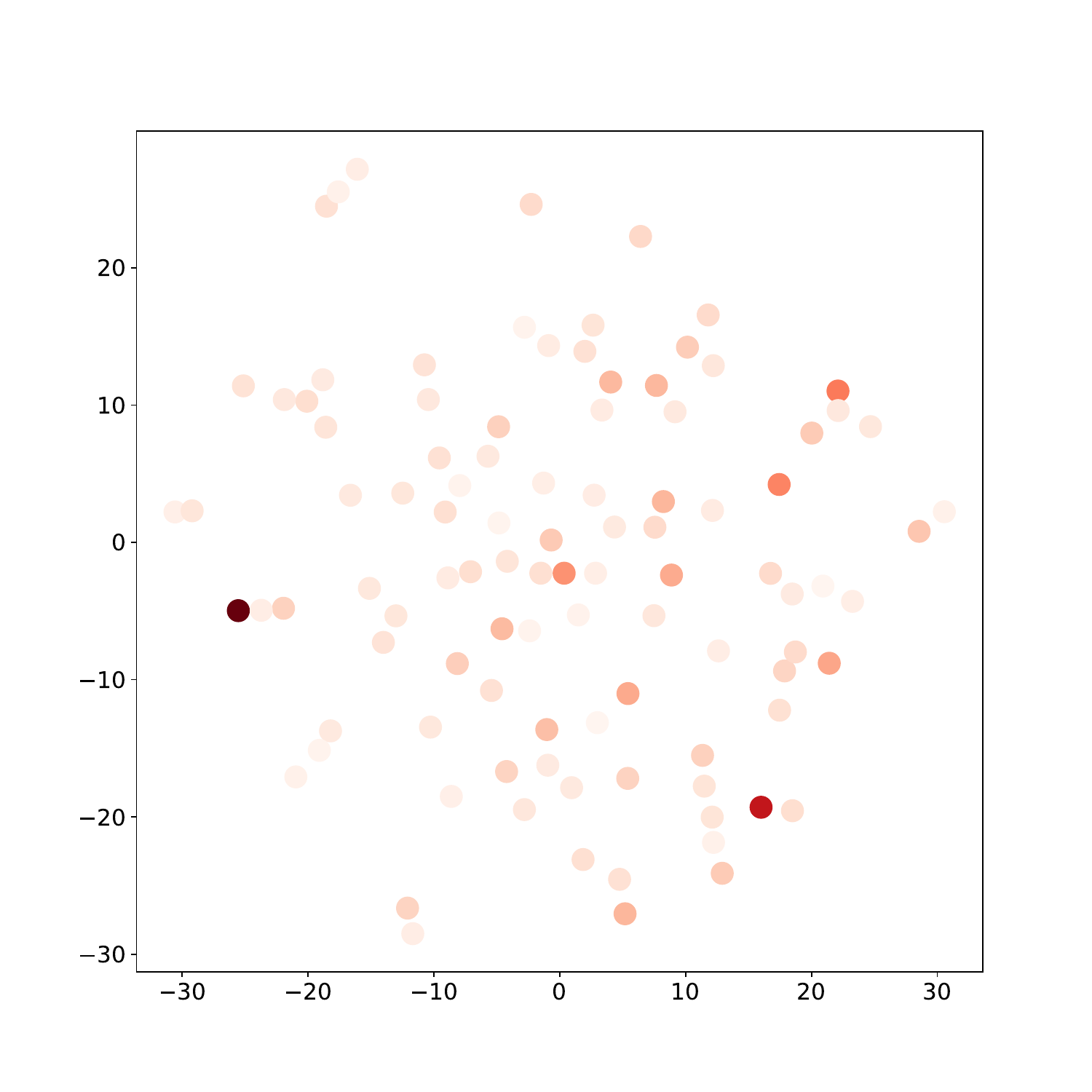}
        \end{subfigure}
    \caption{SBM-a (top) and 
    config (bottom)}
    \end{subfigure}
    \begin{subfigure}[b]{0.5\textwidth}
        \includegraphics[width=1.28\textwidth]{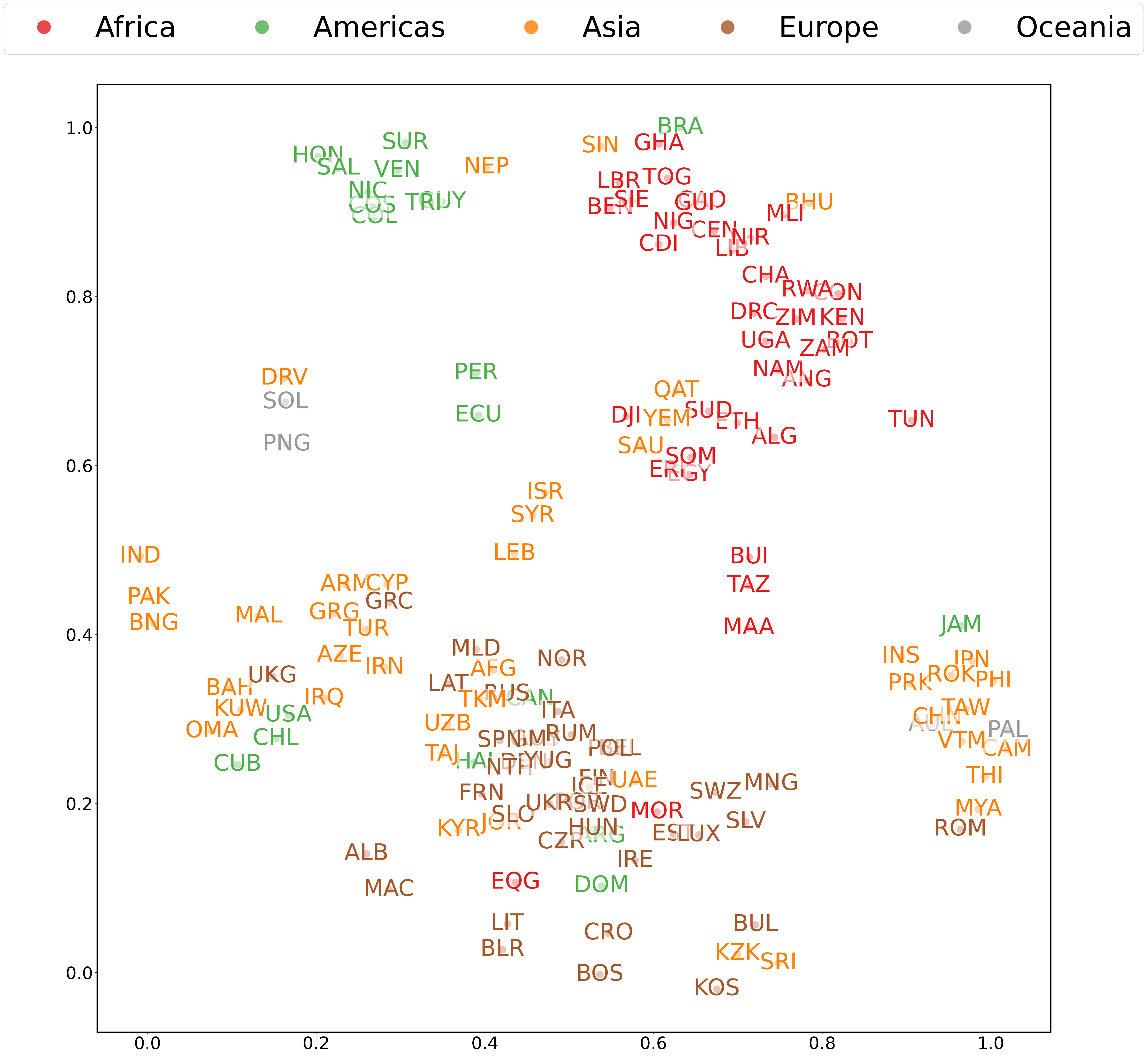}
        \centering
        \caption{MID}
    \end{subfigure}
    \caption{t-SNE visualization of the static node embedding learned by the model. For dynamic SBM, the two colors represent the communities. The opacity of the nodes represents their degree: from lower (light) to higher (dark).}
    \label{fig:embedding}
\end{figure}

\noindent\textbf{Analysis of the base rate. }Figure \ref{fig:embedding} shows the node embeddings learned from the base rate, which are used to infer edge embeddings. For the SBM-a, the embeddings reflect the two original clusters and for config we can see disparities in the degrees of nodes (with opacity of the nodes). For the MID dataset, relevant historical conflicts are in evidence. For example, Syria, Lebanon, and Israel, with their history of enduring conflicts, cluster closely. Similarly, Peru (PER) and Ecuador (ECU), with a long-standing territorial dispute, are also close together. The United States (USA) appears near Cuba (CUB) and Iraq (IRQ), reflecting two significant events: the Cuban Missile Crisis during the Cold War and the U.S.-led invasion of Iraq in 2003. Additionally, the embedding space captures indirect relations (e.g., alliances shaped by shared adversaries) and similarities in conflict profiles, such as countries with low aggression levels compared to others.



\subsubsection{Time Complexity Analysis}

In terms of training time (material details in Appendix \ref{app:mat}), \us demonstrates a significantly lower computational cost compared to both DLS and LSH.
It achieves a minimum speedup of 7 times over DLS, with gains reaching up to 280 times on the Enron dataset.
Similarly, when compared to LSH, \us exhibits a significant efficiency advantage, being 2 to 20 times faster.

All methods share the latent space dimension as a common hyperparameter. In addition, LSH requires three decay parameters, DLS includes the number of kernels, while our model requires no extra hyperparameters beyond the latent space dimension.

 \begin{table}[htbp]
     \centering
     \begin{tabular}{lccccccc}
     \toprule
        Model/Network  & Config.& Stream & SBM-a & SBM-b & Enron & MID& Reality \\
        \midrule
        \us & 7 & 2 & 15 & 19 & 22 & 1087 &  324\\
        \textsc{LSH}&NA&NA&NA&NA&566&2588&715\\
        \textsc{DLS}&417&238&1174&118& 6179 & 7450 & 2092\\
     \bottomrule
     \end{tabular}
     \caption{Training time in seconds. NA stands for Not Applicable.}
     \label{tab:training-time}
 \end{table}
\section{Conclusion and Perspectives}\label{sec:conclusion}
We introduced \framework, a generative framework for temporal networks based on a modified Hawkes Process-based model coupled with a thinning simulation algorithm. Focusing on the inter-arrival times, our model is a simple yet elegant way of capturing the influence of the temporal network topology on the intensity of the dyadic point processes. Our proposed framework suggests that by encoding the history in an appropriate vector of sufficient statistics, one can very simply include inductive bias on the time-evolution of the network, and use the resulting model for simulation. We empirically demonstrated that \us\ can capture and reproduce the dynamic of a network on a wide range of graphs. It balances performance and efficiency, delivering results comparable to or better than state-of-the-art methods. Its ability to excel in specific scenarios while being significantly faster—by several orders of magnitude—makes it a practical tool for graph generation.

As stated before, this work is a first step towards establishing a flexible dynamic graph generation framework, but its limitations are mostly computational. In further work, we plan to integrate representation learning as a substitute for the hand-crafted features \citep{WangCLL021}. This will avoid the problem of choosing which features to compute, and limit the cost of updating them. In addition, it will let us study the effects of higher-order history dependency on the quality of the model. 

\backmatter





\bmhead{Acknowledgements}

The research leading to these results has received funding from the Special Research Fund (BOF) of Ghent University (BOF20/IBF/117), from the Flemish Government under the ``Onderzoeksprogramma Artificiële Intelligentie (AI) Vlaanderen'' programme, and from the FWO (project no. G0F9816N, 3G042220, G073924N). This project has also received funding from the PEPR-IA through the project FOUNDRY.






\bibliography{_main}
\newpage
\begin{appendices}

\section{Algorithmic details of \us.}
We provide the full pseudo-code of the \us~ simulation algorithm.

\begin{algorithm}[H]
\caption{\us~ simulation algorithm}
\label{alg:ogata}
    \begin{algorithmic}
    \State {\bfseries Input:} initial time $t_{n}$, upper bound $T$, max iterations $N$ \\
    \State {\bfseries Initialize:} $t = t_n$, $\mathcal{H}_{t} = \mathcal{H}_{t_n}$, $\lambda^{(*)} = \lambda(t_n | \mathcal{H}_{t_n})$
    
    \While{$t < T$ or $n < N$}
        \State $u \sim \text{Uniform}([0,1])$
        \State $\Delta_t = \min\limits_{e}\left(- \frac{\ln(u^{(e)})}{\lambda^{(e)(*)}}\right)$
        \State Set $t = t + \Delta_t$
    
         \State $\lambda_{\text{current}} = \lambda(t | \mathcal{H}_{t})$ 
        
        \State $U \sim \text{Uniform}([0, \lambda^{(*)}])$ 
        
        \If{$U < \lambda_{\text{current}}$ }
            \State Define for each edge $e \in \mathcal{E}_{\text{active}}$ activated the probability 
            $$p^{(e)} = \frac{\lambda^{(e)}(t | \mathcal{H}_{t})}{\sum_{e' \in \mathcal{E}_{\text{active}}}\lambda^{(e')}(t | \mathcal{H}_{t})}$$
    
            \State Sample $e_{\text{new}} \sim \text{Multinomial}(1, \{ p_e | e \in \epsilon \})$
            
    
            \State Add the event $(e_{\text{new}}, t)$ in $\mathcal{H}_{t}$
    
            \State $t_{last} = t \text{ and } \mathcal{H}_{last} = \mathcal{H}_t $
            \State $\lambda^{(*)} = \lambda(t_{last} | \mathcal{H}_{\text{last}})$
        \EndIf
        \State Set $n = n + 1$
    \EndWhile
    \end{algorithmic}
\end{algorithm}

\section{Additional details and results}\label{app:properties}

\subsubsection{Synthetic Graph Generation}\label{ref:SGG}
Hereafter, we propose to derive different dynamic synthetic models, two of them being dedicated to the generation of static graphs, namely the configuration model and the Stochastic Block Model (SBM), and the third one based on stream graphs \citep{Latapy2017}. For static models, we describe below the modifications made to obtain dynamic versions. The aim is to cover a large diversity of network generation models as the literature tends to focus on SBM models only, therefore assuming a strong community structure of the network. \\

\noindent\textbf{Dynamic Configuration Model.}
The configuration model is a technique used to generate random static networks based on a given degree sequence. Frequently used as a reference for real social networks, it allows the integration of various degree distributions. We take advantage of this model to create dynamic graphs, excluding self-loops.
The procedure is defined as follows. We first choose a fixed number of node $\{v_1,..., v_n\}$ with their associated degree sequence $[d_1,...,d_n]$ for $ i \in \{1,...,n\}$,  $d_i  \sim \mathcal{N}(m_i, \sigma)$ and generate our initial graph based on the configuration model. Then, for a fixed number of iterations, we repeat the following two steps: (1) modify $m_i$ by incrementing a predefined natural integer drawn from a uniform distribution and simulate the associated graph; (2) add the edge list from the previous graph, including associated timestamps, into the final simulation.\\

\noindent\textbf{Temporal SBM.}
Secondly, we use a generative process based on SBM. We fix two communities, similar to the configuration model we set a fixed number of iterations. At each iteration, we generate an SBM graph with a fixed block probability and add random timestamps to each formed edge. The final SBM multigraph is the concatenation of the different SBM graphs along with their timestamps. We also create a scenario where one of the communities disappears by merging with the other. After some time the unique community gets split again.\\

\noindent\textbf{Stream Graph}
We opt for the stream graph approach \citep{Latapy2017} to create the last type of graph. Stream graphs depict temporal data sequences alongside nodes and links, providing a dynamic perspective on evolving node relationships over time. 
We use \texttt{Straph}\footnote{https://github.com/StraphX/Straph?tab=readme-ov-file} that simulates stream graphs that follow Erd\H{o}s-R\'enyi behavior (i.e., the activation of nodes follows a Poisson distribution). The time associated with each edge activation in the stream graph is continuous, meaning that each edge has a time interval of activation instead of a discrete time as in our framework. To convert time interval into discrete time we select $n$ discrete times within this interval, where $n$ is proportional to the length of the interval (i.e., for a longer interval, we will have more edge activations than for a shorter interval).


\subsection{Network properties}\label{ref:NETprop}

Table \ref{tab:dataset} presents additional details regarding the synthetic and real-world networks.
\begin{table}[h]
	\begin{center}
		\resizebox{0.7\textwidth}{!}{
				\begin{tabular}{lcccc}
					\toprule
					\textbf{Dataset}    & \textbf{\#events} &   \textbf{\#nodes}   & \textbf{Density} &   \textbf{CC} \\
					 \midrule
                    Config. & $ 2800 \pm 75$& 100& $0.234 \pm 0.004$ & $0.31\pm0.01$ \\
                    Stream & $2307 \pm 88 $& 100& $ 0.101\pm0.003$  & $0.1\pm0.01$\\
                    SBM-a& $1956\pm 26$& 100&  $0.288\pm 0.005 $& $ 0.5\pm 0.01$\\
                    SBM-b & $ 2791 \pm 32$& 100&$ 0.385 \pm 0.005$  &$ 0.6\pm 0.01 $ \\
					Enron & 4529& 167& 0.31  & 0.06\\
					MID & 5088 & 145 & 0.03& 0.27\\
                    Reality & 2159 & 92 & 0.04 & 0.24  \\
                    \bottomrule
				\end{tabular}
		
		}
	\end{center}
	\caption{Properties of the synthetic and real-world networks. We report the number of events that occurred during the time of observation, the number of nodes, the density of the network, and the clustering coefficient (CC).  For simulated networks, we report mean $\pm$ std computed over 10 synthetic networks generated with the same configuration.}
	\label{tab:dataset}
\end{table}

\subsection{Degree centrality nodes}
\label{app:clustering_coef}
\begin{figure}[h]
    \centering
    \begin{subfigure}[b]{\textwidth}
        \centering
        \includegraphics[width=1\textwidth]{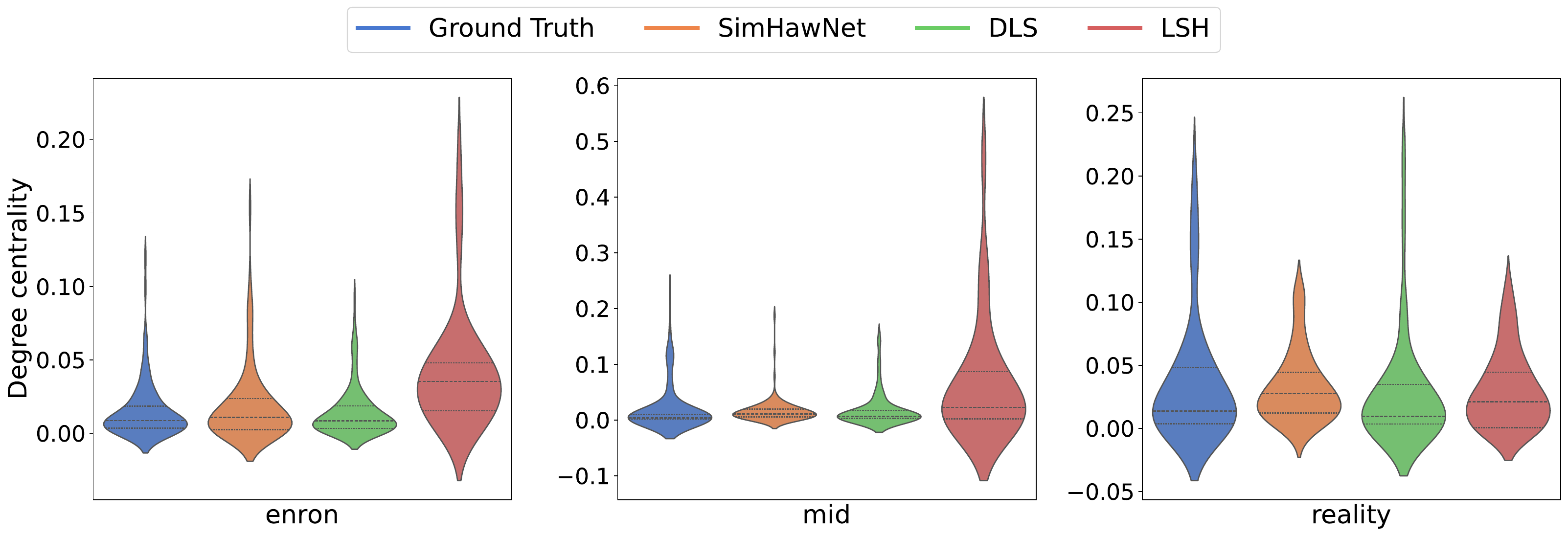} 
        \label{fig:image1}
    \end{subfigure}

    \vspace{1cm} 

    \begin{subfigure}[b]{\textwidth}
        \centering
        \includegraphics[width=1\textwidth]{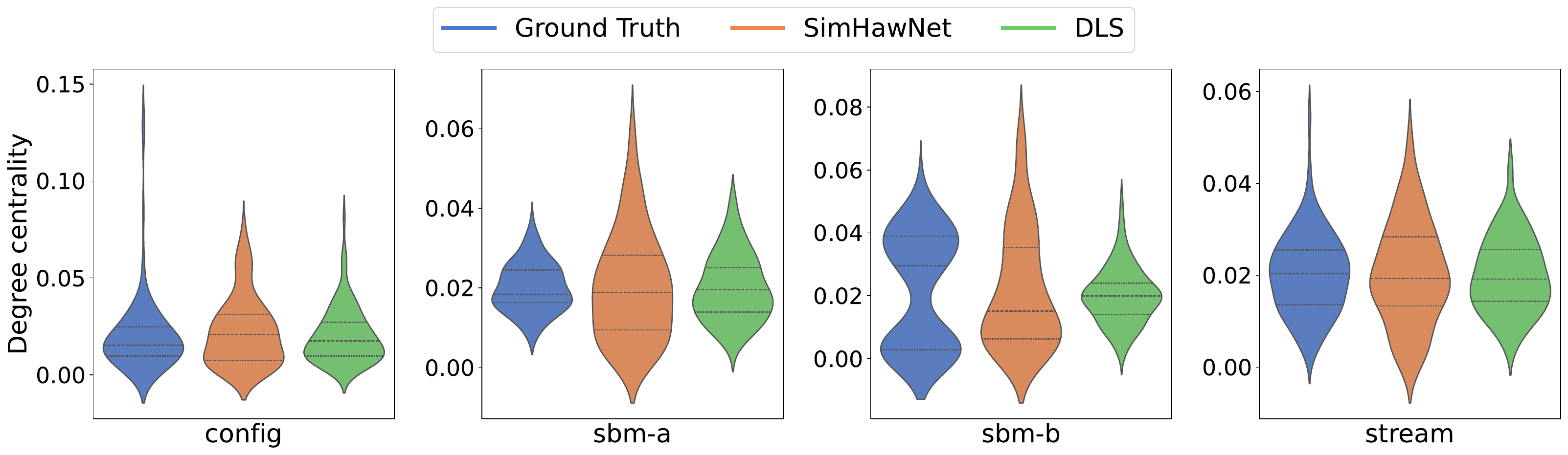} 
        \label{fig:image2}
    \end{subfigure}

    \caption{Distribution of node degree centrality}
    \label{fig:figure_degree_centrality}
\end{figure}

Figure \ref{fig:figure_degree_centrality} illustrates the distribution of degree centrality between nodes in graph simulations performed by different models: \us, DLS and LSH. Our model demonstrates superior performance in capturing node centrality for the SBM-b case compared to DLS. This dataset represents a graph undergoing dynamic transformations over time, with clusters activated sequentially.

Furthermore, we observe that our model identifies central nodes more efficiently. This is evident in datasets such as Enron, MID, SBM-b and Stream, where our model produces more elongated violin plots covering a range closely corresponding to the ground truth. This indicates better alignment with the true distribution of degree centrality.

\subsection{Global Intensity Results}
\label{app:intensities}
\begin{figure}[h]
    \centering
    \begin{subfigure}[b]{0.49\textwidth}
        \centering
        \includegraphics[width=\textwidth]{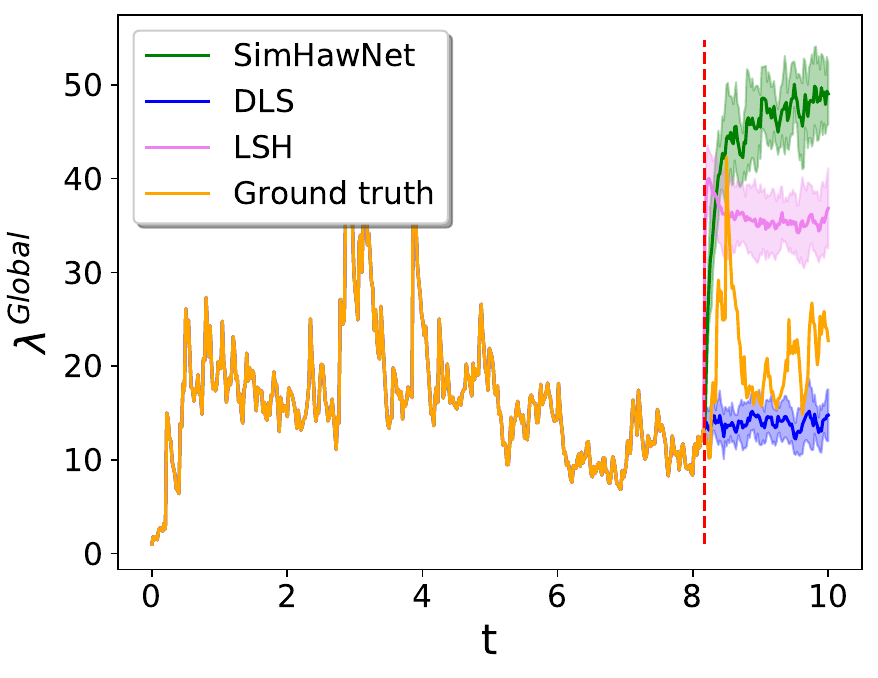}
        \caption{MID}
        \label{fig:mid}
    \end{subfigure}
    \begin{subfigure}[b]{0.49\textwidth}
        \centering
        \includegraphics[width=\textwidth]{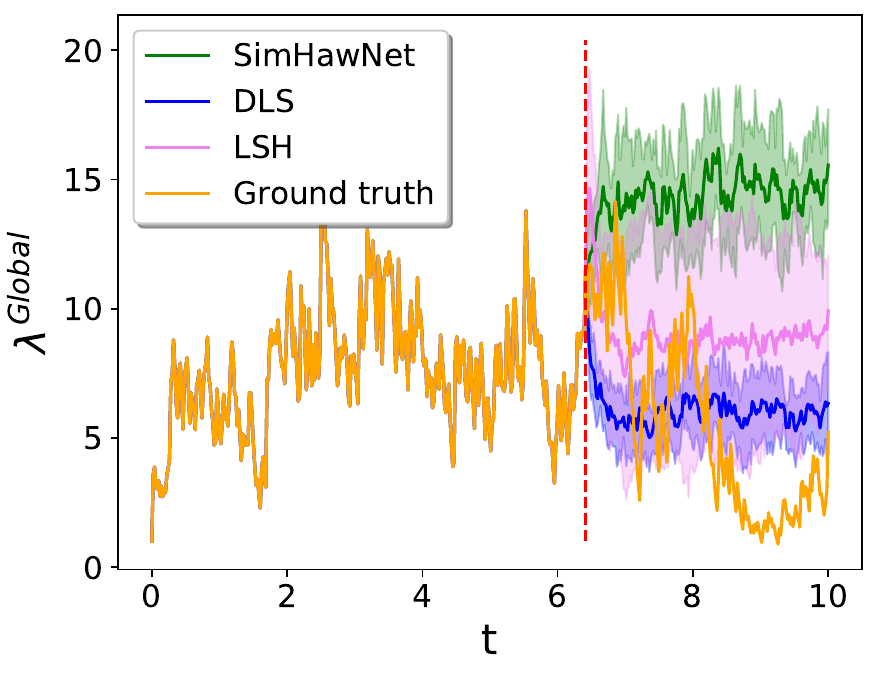}
        \caption{Reality}
        \label{fig:reality}
    \end{subfigure}

    \begin{subfigure}[b]{0.32\textwidth}
        \centering
        \includegraphics[width=\textwidth]{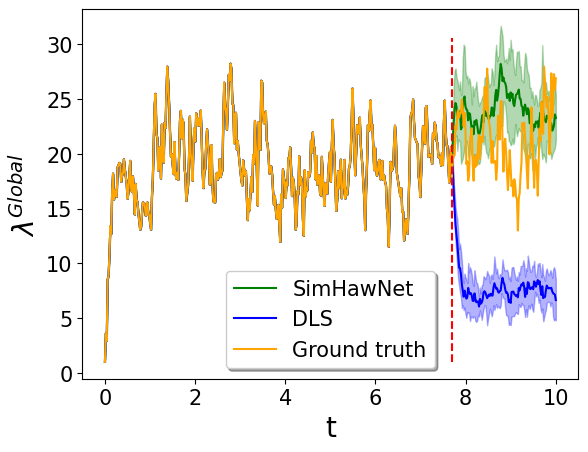}
        \caption{sbm-a}
        \label{fig:sbm_a}
    \end{subfigure}
    \begin{subfigure}[b]{0.32\textwidth}
        \centering
        \includegraphics[width=\textwidth]{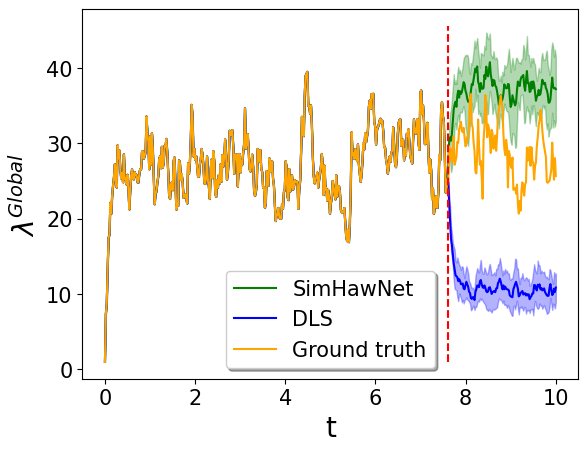}
        \caption{sbm-b}
        \label{fig:sbm_b}
    \end{subfigure}
    \begin{subfigure}[b]{0.32\textwidth}
        \centering
        \includegraphics[width=\textwidth]{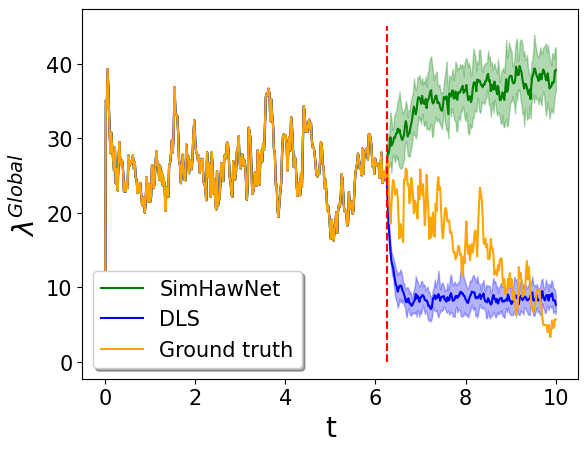}
        \caption{stream}
        \label{fig:stream}
    \end{subfigure}

    \caption{Global intensity functions obtained. The red line marks the beginning of the simulation. Each subfigure corresponds to different datasets or configurations.}
    \label{fig:intensity_combined_2}
\end{figure}

Figure \ref{fig:intensity_combined_2} presents the intensities produce by our model, DLS and LSH on the other dataset.

\subsection{Parameters analysis}
\label{app:param}
\begin{figure}[!h] 
    \centering 
   \includegraphics[width=0.98\textwidth]{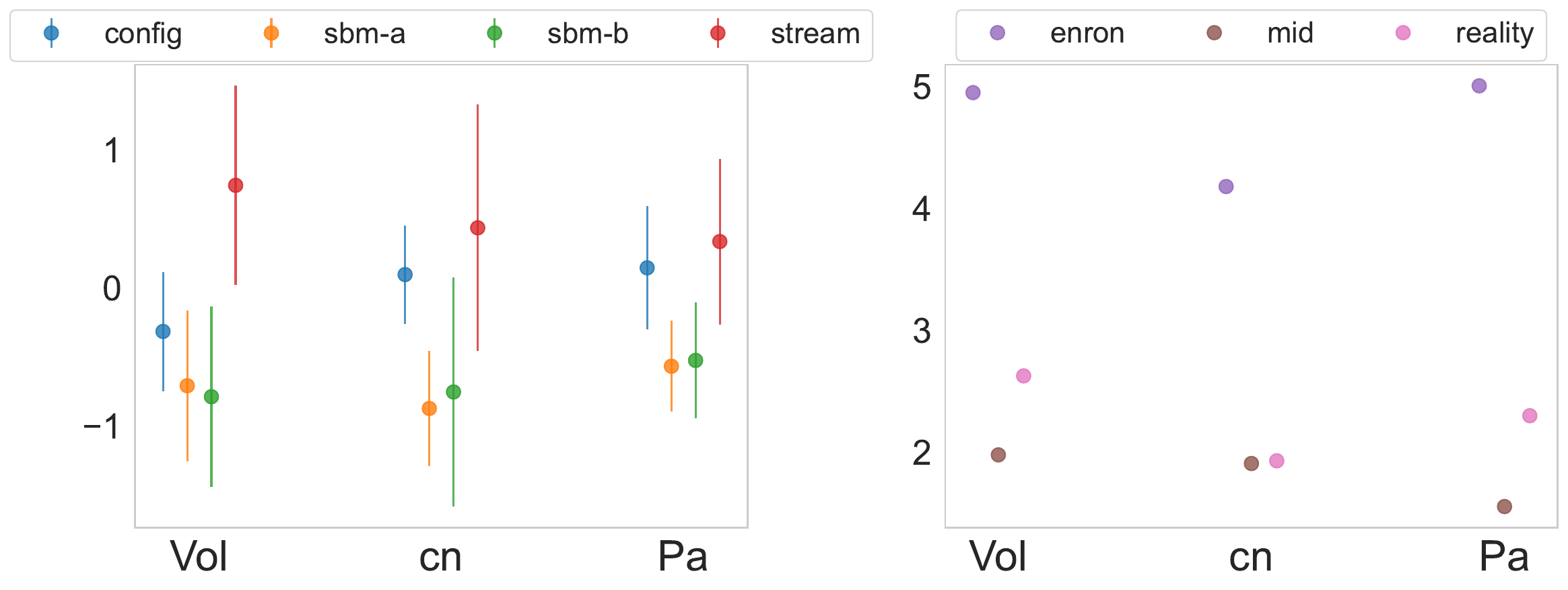}
    \caption{Parameters of $g$ learned by our model for (left) synthetic networks and (right) real-world networks. Features: volume (\textbf{Vol}), common neighbors (\textbf{cn}), and preferential attachment (\textbf{Pa}).}
    \label{fig:param_g} 
\end{figure}

Figure \ref{fig:param_g} shows the mean and variance of the temporal feature weights learned by our model in different graphs. We observe a consistent pattern for the SBM-a, Enron and Reality datasets: the volume and preferential attachment features have similar importance, while the common neighbors feature has a comparatively weaker influence.

\begin{figure}[!h]
    \centering
    \makebox[\textwidth][c]{%
        \includegraphics[width=1.\textwidth]{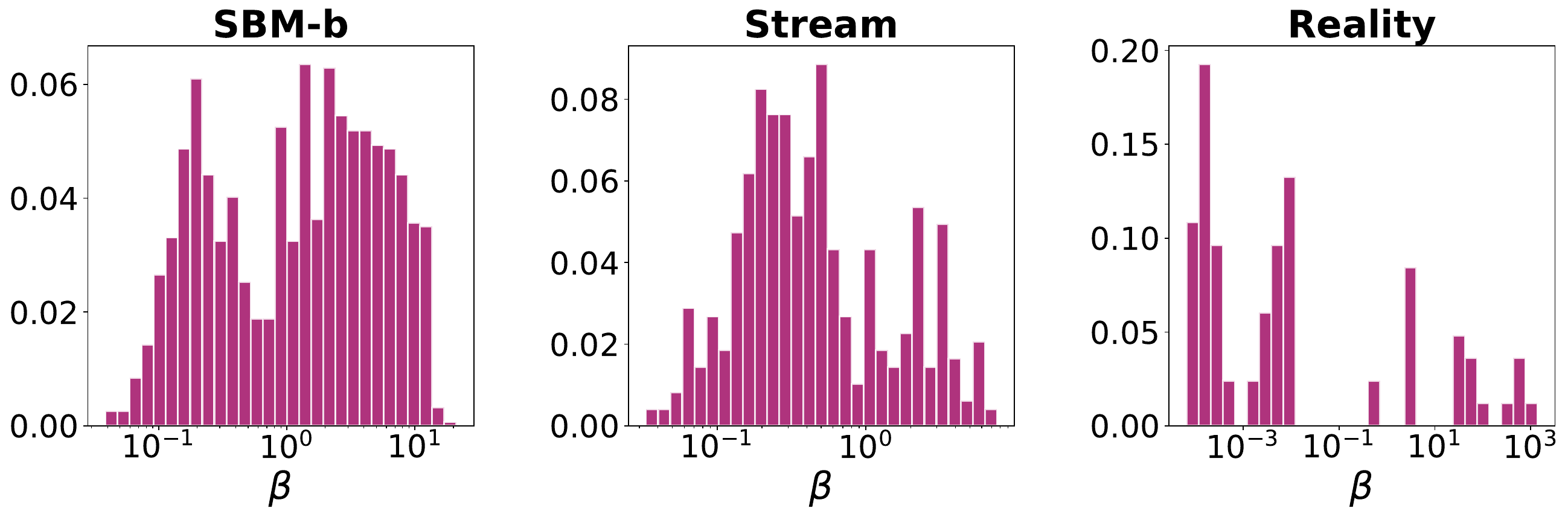}
    }
    \caption{Plot the percentage of the decay ($\beta$) values across the edges for the model on different learned datasets, with a logarithmic scale on the x-axis for $\beta$.}
    \label{fig:beta_distribution_app}
\end{figure}

Figure \ref{fig:beta_distribution_app}  shows the distribution of the $\beta$ parameters learned by the model. 

\subsection{Material Details}
\label{app:mat}
We train the models on 2x Intel Xeon E5-2695 CPUs.

\end{appendices}




\end{document}